\documentclass[10pt,journal,compsoc]{IEEEtran}
\usepackage{times}
\usepackage{epsfig}
\usepackage{graphicx}
\usepackage{subfigure}
\usepackage{amsmath}
\usepackage{amssymb}
\usepackage{hyperref}
\usepackage{color}
\usepackage{multirow}
\usepackage{pgfplots}
\usepackage{adjustbox}
\usepackage{amsfonts}
\usepackage{tikz}
\usepackage{enumitem}

\definecolor{ao(english)}{rgb}{0.0, 0.5, 0.0}

\newcommand{\Chen}[1]{\textcolor{red}{Chen: #1}}
\newcommand{\rui}[1]{\textcolor{blue}{Rui says: #1}}

\ifCLASSOPTIONcompsoc
  \usepackage[nocompress]{cite}
\else
  \usepackage{cite}
\fi

\ifCLASSINFOpdf
\else
\fi

\pgfplotsset{compat=1.14}

\begin{document}
\title{An End-to-end 3D Convolutional Neural Network for Action Detection and Segmentation in Videos}

\author{Rui~Hou,~\IEEEmembership{Student Member,~IEEE,}
        Chen~Chen,~\IEEEmembership{Member,~IEEE,}
        and~Mubarak~Shah,~\IEEEmembership{Fellow,~IEEE}
\IEEEcompsocitemizethanks{\IEEEcompsocthanksitem The authors are with the Center for Research in Computer Vision, University of Central Florida, Orlando, FL, 32816. E-mail: houray@gmail.com; chenchen870713@gmail.com; shah@crcv.ucf.edu \protect\\}

}

\markboth{Journal of \LaTeX\ Class Files,~Vol.~14, No.~8, August~2015}%
{Shell \MakeLowercase{\textit{et al.}}: Bare Demo of IEEEtran.cls for Computer Society Journals}

\IEEEtitleabstractindextext{%
\begin{abstract}
Deep learning has been demonstrated to achieve excellent results for image classification and object detection. However, the impact of deep learning on video analysis (\eg~action detection and recognition) has not been that significant due to complexity of video data and lack of annotations. In addition, training deep neural networks on large scale video datasets is extremely computationally expensive. Previous convolutional neural networks (CNNs) based video action detection approaches usually consist of two major steps: frame-level action proposal generation and association of proposals across frames. Also, most of these methods employ two-stream CNN framework to handle spatial and temporal features separately.  In this paper, we propose an end-to-end 3D CNN for action detection and segmentation in videos. The proposed architecture is a unified deep network that is able to recognize and localize action based on 3D convolution features. A video is first divided into equal length clips and next for each clip a set of tube proposals are generated  based on 3D CNN features. Finally, the tube proposals of different clips are linked together and spatio-temporal action detection is performed using these linked video proposals. This top-down action detection approach explicitly relies on a set of good tube proposals to perform well and training the bounding box regression usually requires a large number of annotated samples. To remedy this, we further extend the 3D CNN to an encoder-decoder structure and formulate the localization problem as action segmentation. The  foreground regions (\ie~action regions) for each frame  are  segmented  first then  the segmented foreground
maps  are used  to generate the  bounding boxes. This bottom-up approach effectively avoids tube proposal generation by leveraging the pixel-wise annotations of segmentation. The segmentation framework also can be readily applied to a general problem of video object segmentation. Extensive experiments on several video datasets demonstrate the superior performance of our approach for action detection and video object segmentation compared to the state-of-the-arts.
\end{abstract}

}

\maketitle
\IEEEdisplaynontitleabstractindextext
\IEEEpeerreviewmaketitle

\IEEEraisesectionheading{
\section{Introduction}
\label{sec:introduction}
}
\IEEEPARstart{T}{he} goal of action detection is to detect every occurrence of a given action within a long video, and to localize each detection both in space and time. Deep learning learning based approaches have significantly improved video action recognition performance. Compared to action {\em recognition}, action {\em detection} is a more challenging task due to flexible volume shape and large spatio-temporal search space.

Previous deep learning based action detection approaches first detect frame-level action proposals by popular proposal algorithms \cite{gkioxari2015finding,weinzaepfel2015learning} or by training proposal networks \cite{peng2016multi}. Then the frame-level action proposals are associated across frames to determine final action detection through tracking based approaches. Moreover, in order to capture both spatial and temporal information of an action, two-stream networks (a spatial CNN and an optical flow CNN) are used. In this manner, the spatial and motion information are processed separately.

Region Convolution Neural Network (R-CNN) for object detection in images was proposed by Girshick \etal \cite{rcnn_Girshick_2014_CVPR}. It was followed by a fast R-CNN proposed in \cite{fast_rcnn_Girshick_2015_ICCV}, which includes the classifier as well. Later, Faster R-CNN \cite{faster_rcnn} was developed by introducing a region proposal network. It has been extensively used to produce excellent results for object detection in images.
A natural generalization of the R-CNN from 2D images to 3D spatio-temporal volumes is to study their effectiveness for the problem of action detection in videos. A straightforward spatio-temporal generalization of the R-CNN approach would be to treat action detection in videos as a set of 2D image detection using Faster R-CNN. However, unfortunately, this approach does not take the temporal information into account and is not sufficiently expressive to distinguish between actions.

Inspired by Faster R-CNN \cite{faster_rcnn}, we propose Tube Convolutional Neural Network (T-CNN) for action detection by leveraging the descriptive power of 3D CNN. To better capture the spatio-temporal information of video, we exploit 3D CNN since it is able to capture motion characteristics in videos and shows promising results on video action recognition. In our approach, an input video is divided into equal length clips first. Then, the clips are fed into Tube Proposal Network (TPN) and a set of tube proposals are obtained. Next, tube proposals from each video clip are linked according to their actionness scores and overlap between adjacent proposals to form a complete tube proposal for spatio-temporal action localization in the video. Finally, the Tube-of-Interest (ToI) pooling is applied to the linked action tube proposal to generate a fixed size feature vector for action label prediction.

Our T-CNN is generalization of Faster R-CNN. However, Faster R-CNN relies on a set of anchor boxes with different sizes and aspect ratios, and applies them in a sliding window fashion for object detection. Bounding box regression is used to refine the anchor boxes in terms of position and size for setting a tight boundary around the object. Such object detection methods are considered as top-down detectors, which are similar in spirit to the top-down, class-specific segmentation algorithms \cite{borenstein2002class,winn2005locus,tu2005image} which use the shape of the deformed model of a known object to estimate the desired segmentation.
Although top-down detection methods have achieved promising results, they still face several issues. First, bounding boxes only provide coarse localization information as compared to pixel-wise segmentation map. Second, to train bounding box regression, a large amount of training data is necessary. Also, since human actions are complex with dynamic visual patterns, it would require variable templates for each fame of a video for top-down algorithms, leading to the challenge of preparing a set of good anchor boxes priors.

\begin{figure}
	\centering
	\includegraphics[width=0.95\linewidth]{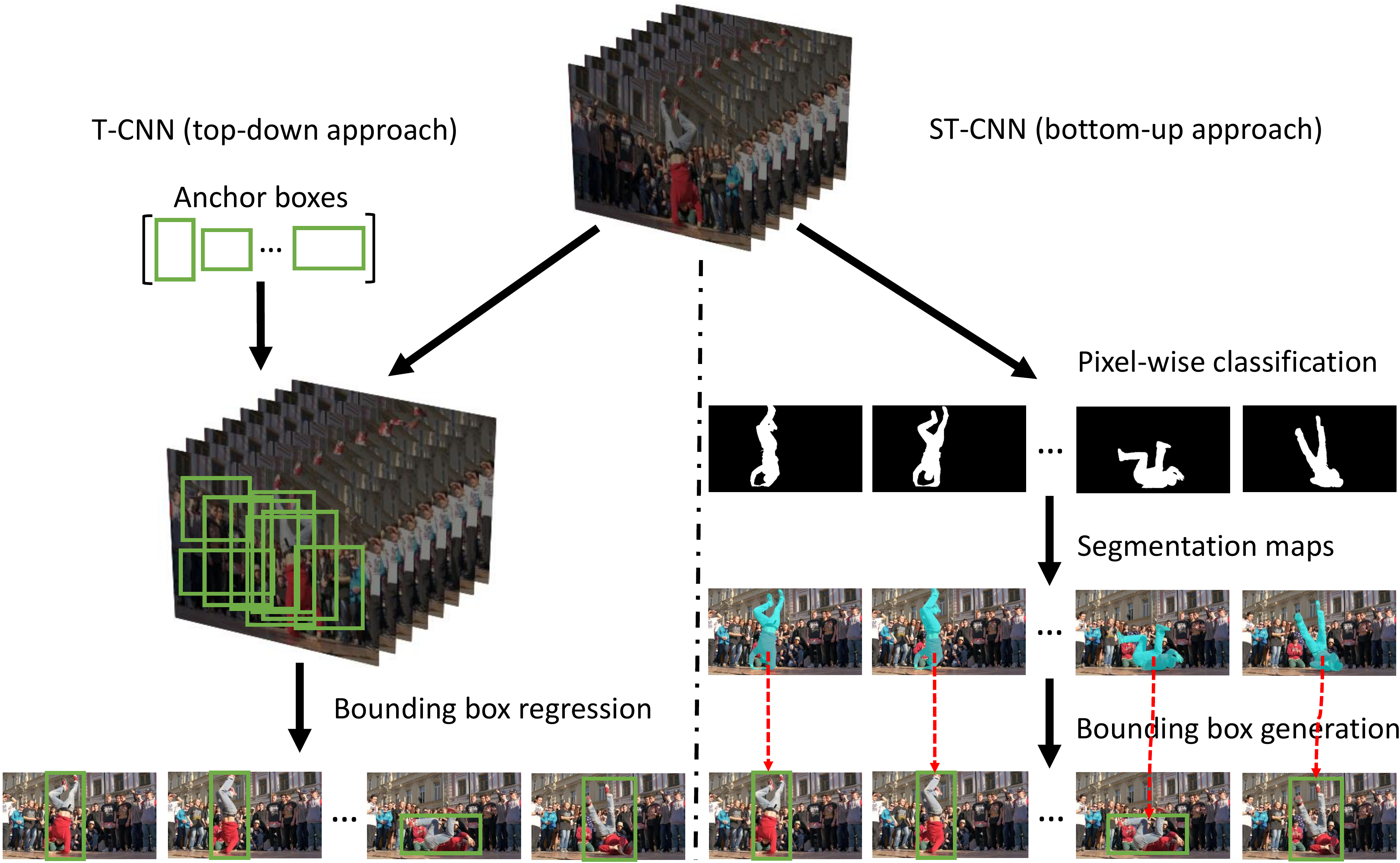}
	\caption{Top-Down (T-CNN) and Bottom-Up  (ST-CNN) approaches. The top-down  approach takes the whole frame as the search space with some predefined bounding box templates, filters candidate bounding box proposals based on their confidence scores, and performs bounding box regression on the selected proposals. While, the Bottom-Up  approach directly operates on pixels. It takes only segments within frames as candidates and agglomerates them as bounding box detection.}
	\label{fig:mtcnn-Teaser}
\end{figure}

Low-level features like intensity, orientation, and motion can render measurement of actionness for every pixel in image.
Therefore, we also propose to explore an alternative strategy for action detection using bottom-up action segmentation. Assuming pixel-wise annotations (action foreground and background) are available, an encoder-decoder network structure, which is commonly used for image semantic segmentation \cite{badrinarayanan2015segnet}, using 3D CNN is developed to predict pixel-wise labels for each video frame.
Then the pixel-wise predictions (\ie~action segmentation map) for each frame are used to infer the corresponding bounding boxes by finding the boundary of the foreground map. After bounding box generation, ToI pooling is applied to feature tubes for action recognition. We dub the segmentation based approach as Segmentation T-CNN (ST-CNN).
The general frameworks of T-CNN and ST-CNN are compared in Figure~\ref{fig:mtcnn-Teaser}. The main difference between ST-CNN and T-CNN is that ST-CNN avoids generating tube proposals by treating action detection as a binary (\ie~foreground action and background) video segmentation task.

To the best of our knowledge, this is the first work to exploit 3D CNN for video action detection (\ie~localization and recognition) in an end-to-end fashion. As an extension to our ICCV paper \cite{hou2017tube}, the main contributions of this paper are summarized as follows. First, we improve the T-CNN framework by proposing a bottom-up approach, ST-CNN, which leverages pixel-wise action segmentation for action detection.
Second, we also use ST-CNN for video object segmentation and show competitive results comparing to the state-of-the-arts.
Third, we conduct a thorough comparison between T-CNN and ST-CNN on action detection, and provide insights in terms of performance and generalization ability.
Moreover, through extensive experiments, we demonstrate that our end-to-end 3D CNN approach has the capacity to achieve superior performance over the popular two-stream network pipeline for video analysis tasks, \eg~action detection and video object segmentation.

The remainder of this paper is organized as follows. Section \ref{sec:related_work} reviews related work on action recognition and detection. Section \ref{sec:difference} discusses the challenges of generalizing R-CNN from 2D to 3D. Section \ref{sec:approach} provides detailed procedures of the proposed T-CNN. In Section \ref{sec:seg_overview}, we introduce the bottom-up approach, ST-CNN, for action detection and segmentation.
The experimental results on action detection and video object segmentation are presented in Section \ref{sec:experiments}.
Finally, Section \ref{sec:conclusion} concludes the paper.

\section{Related Work}\label{sec:related_work}

\textbf{Object detection.} For object detection in images, Girshick \etal~propose Region CNN (R-CNN) \cite{rcnn_Girshick_2014_CVPR}.  In their approach region proposals are extracted using selective search. Then the candidate regions are warped to a fixed size and fed into CNN to extract CNN features. Finally, SVM model is trained for object classification. A fast version of R-CNN, Fast R-CNN, is presented in \cite{fast_rcnn_Girshick_2015_ICCV}. Compared to the multi-stage pipeline of R-CNN, fast R-CNN incorporates object classifier in the network and trains object classifier and bounding box regressor simultaneously. Region of interest (RoI) pooling layer is introduced to extract fixed-length feature vectors for bounding boxes with different sizes. After that, Faster R-CNN \cite{faster_rcnn} introduces a Region Proposal Network (RPN) to replace selective search for proposal generation. RPN shares full image convolutional features with the detection network, thus the proposal generation is almost cost-free. A recent extension of Faster R-CNN is the Mask R-CNN \cite{he2017mask}, which adds another branch for predicting object mask within each bounding box detection. SSD (Single Shot MultiBox Detector) \cite{liu2016ssd} and YOLO (you only look once) \cite{redmon2016you,yolo9000} detectors eliminate proposal generation and detect objects in a single network, thus greatly improving the detection speed.
These state-of-the-art object detectors are served as the basis for recent action detection methods~\cite{peng2016multi,singh2016online,kalogeiton2017joint}.

\textbf{3D CNN.} Convolutional neural networks have been demonstrated to achieve excellent results for action recognition \cite{lecun2015deep}. Karpathy \etal \cite{karpathy2014large}  explore various frame-level fusion methods over time. Ng \etal \cite{lstm_ng} use recurrent neural network employing the CNN feature. Since these approaches only use frame based CNN features, the temporal information is neglected. Simonyan \etal \cite{2stream_cnn_simonyan_2014two} propose the two-stream CNN approach for action recognition. Besides a classic CNN which takes images as an input, it has a separate network for optical flow. Moreover, Wang \etal~fuse the trajectories and CNN features. Although these methods, which take hand-crafted temporal feature as a separate stream, show promising performance on action recognition, they do not employ end to end deep network and require separate computation of  optical flow  and optimization of the parameters. 3D CNN is a logical solution to this issue. Ji \etal \cite{ji20133d} propose a 3D CNN based human detector and head tracker to segment human subjects in videos. Sun \etal \cite{sun2015human} propose a factorization of 3D CNN and exploit multiple ways to decompose convolutional kernels. Tran \etal \cite{c3d} use 3D CNN for large scale action recognition. Leveraging the 3D CNN model \cite{c3d} trained on Sport1M video dataset \cite{karpathy2014large}, the 3D CNN framework has been exploited for temporal action localization in untrimmed videos \cite{shou2016temporal,shou2017cdc,xu2017r,sstad_buch_bmvc17,buch2017sst}.
However, to the best of our knowledge, we are the first to exploit 3D CNN for action {\em detection}.

\textbf{Action detection.} Compared to action recognition, action detection is a more challenging problem \cite{jain201515}, which has been an active area of research. Before the deep learning era, the top-down based approaches dominated the topic first. Ke \etal \cite{ke2007event} present an approach for event detection in crowded videos by matching event models to over-segmented spatio-temporal volumes. Lan \etal \cite{lan2011discriminative} treat the spatial location of human in the video as a latent variable, and perform recognition and detection simultaneously. Tian \etal \cite{tian2013spatiotemporal} develop spatio-temporal deformable parts model \cite{dpm} to detect actions in videos. Oneata \etal \cite{oneata2014efficient} and Desai \etal \cite{desai2012detecting} propose sliding-windows based approaches.


Another line of research explores the bottom-up strategy. Bottom-up based approaches replace the exhaustive search, which is a common strategy in top-down  approaches, with selective search. The selective search has a much smaller search space than the exhaustive one. In other words, selective search based approaches take less time to perform ``search'' in frames. Therefore, with the same processing time, these approaches are able to take advantage of more advanced feature representations such as Fisher vector or VLAD instead of basic HOG and HOF. Ma \etal \cite{ma2013action} use a linear SVM on a bag of hierarchical space-time segments representation. They build a spatio-temporal feature pyramid followed by latent SVM. Jain \etal \cite{jain2014action} expand the 2D super-pixels to 3D super-voxels. Super-voxels are merged according to appearance and motion costs to produce multiple segmentation layers. Authors in \cite{soomro2015action} and \cite{oneata2014spatio} over segment a video into super-voxels, which are merged into action proposals. The proposals are used to determine action labels for  detection. Lu \etal \cite{lu2015human} propose a hierarchical Markov Random Field (MRF) model to connect different levels of the super-voxel hierarchies for action detection.

The success of  deep learning algorithms, particularly CNN, in images paved the way to use them  for action detection in video. Authors in \cite{gkioxari2015finding} extract frame-level action proposals using selective search and link them using Viterbi algorithm. While in \cite{weinzaepfel2015learning} frame-level action proposals are obtained by EdgeBox \cite{zhu2015tracking} and linked by a tracking algorithm.
Motivated by Faster R-CNN \cite{faster_rcnn}, two-streams R-CNNs for action detection is proposed in \cite{peng2016multi,kalogeiton2017joint}, where a spatial RPN and a motion RPN are used to generate frame-level action proposals. Kalogeiton \etal \cite{kalogeiton2017action} extend the spatial only proposal to spatio-temporal proposal by inducing action tubelet. Singh \etal \cite{singh2016online} adopt SSD to perform online spatio-temporal action localization in real-time.
However, these deep learning based approaches
treat the spatial  and temporal features of a video separately by training two-stream 2D CNN. Therefore, the temporal consistency in videos is not well explored in the network. In contrast, in the proposed T-CNN, we determine action tube proposals directly from input videos and extract compact and more effective spatio-temporal features using 3D CNN.

It is also worth noting that the existing deep learning based approaches operate in a top-down fashion, which requires exhaustive search in the whole frame followed by narrowing down to several appropriate bounding boxes. Clearly, a bottom-up based approach is more efficient. We propose a segmentation based action detection approach (ST-CNN), which accumulates foreground pixels together to form segmentation regions in all frames. And we are the first to generalize the encoder-decoder style network from 2D to 3D for action detection and segmentation.

\textbf{Video object segmentation.} Video object segmentation aims to delineate the foreground object(s) from the background in all frames. Several CNN based approaches have been proposed for this task.
Existing methods \cite{cheng2017segflow,jain2017fusionseg,tokmakov2016learning} leverage two-stream pipeline to train a RNN \cite{tokmakov2017learning} model or learn motion patterns for moving object segmentation \cite{tokmakov2016learning}. While taking inspiration from these works, we are the first to present a 3D CNN based deep framework for video object segmentation in a fully automatic manner without human selected feature.

\section{Generalizing R-CNN from 2D to 3D}
\label{sec:difference}
Generalizing R-CNN from 2D image regions to 3D video tubes is challenging due to the asymmetry between space and time.
Different from images which can be cropped and reshaped into a fixed size, videos vary widely in temporal dimension. Therefore, we divide input videos into fixed length (8 frames) clips, so that video clips  can be processed with a fixed-size CNN architecture. Also, clip based processing mitigates the cost of GPU memory.

\begin{table}
\caption{Network architecture of T-CNN. We refer kernel with shape $d\times h \times w$ where $d$ is the kernel depth, $h$ and $w$ are height and width. Output matrix with shape $C\times D\times H\times W$ where $C$ is the number of channels, $D$ is the number of frames (depth), $H$ and $W$ are the height and width of feature maps.
}
\begin{center}
\begin{tabular}{lcc}
\hline
name            & kernel dims        & output dims \\
 &  ($d\times h \times w$) & ($C\times D\times H\times W$) \\
\hline
conv1           & $3\times3\times3$ & $64\times8\times300\times400$ \\
max-pool1       & $1\times2\times2$ & $64\times8\times150\times200$ \\
conv2           & $3\times3\times3$ & $128\times8\times150\times200$ \\
max-pool2       & $2\times2\times2$ & $128\times4\times75\times100$ \\
conv3a          & $3\times3\times3$ & $256\times4\times75\times100$ \\
conv3b          & $3\times3\times3$ & $256\times4\times75\times100$ \\
max-pool3       & $2\times2\times2$ & $256\times2\times38\times50$ \\
conv4a          & $3\times3\times3$ & $512\times2\times38\times50$ \\
conv4b          & $3\times3\times3$ & $512\times2\times38\times50$ \\
max-pool4       & $2\times2\times2$ & $512\times1\times19\times25$ \\
conv5a          & $3\times3\times3$ & $512\times1\times19\times25$ \\
conv5b          & $3\times3\times3$ & $512\times1\times19\times25$ \\
\hline
toi-pool2      & --                & $128\times8\times8\times8$ \\
toi-pool5       & --                & $512\times1\times4\times4$ \\
1x1 conv        & --                & $8192$ \\
\hline
fc6             & --                & $4096$ \\
fc7             & --                & $4096$ \\
\hline
\end{tabular}
\end{center}
\label{tab:architecture}
\end{table}

To better capture the spatio-temporal information in video, we exploit 3D CNN for action proposal generation and action recognition. One advantage of 3D CNN over 2D CNN is that it captures motion information by applying convolution in both time and space. Since 3D convolution and 3D max pooling are utilized in our approach, not only in the spatial dimension but also in the temporal dimension, the size of video clip is reduced while relevant information is captured. As demonstrated in \cite{c3d}, the temporal pooling is important for recognition task since it better models the spatio-temporal information of video and reduces background noise. However, the temporal order is lost in the process. That means if we arbitrarily change the order of the frames in a video clip, the resulting 3D max-pooled feature cube will be the same. This is problematic in action {\em detection}, since it relies on the feature cube to get bounding boxes for the original frames. To this end, we incorporate temporal skip pooling to retain temporal order residing in the original frames. More details are provided in the next section.

\begin{figure}[!htb]
\centering
\includegraphics[width=0.95\linewidth]{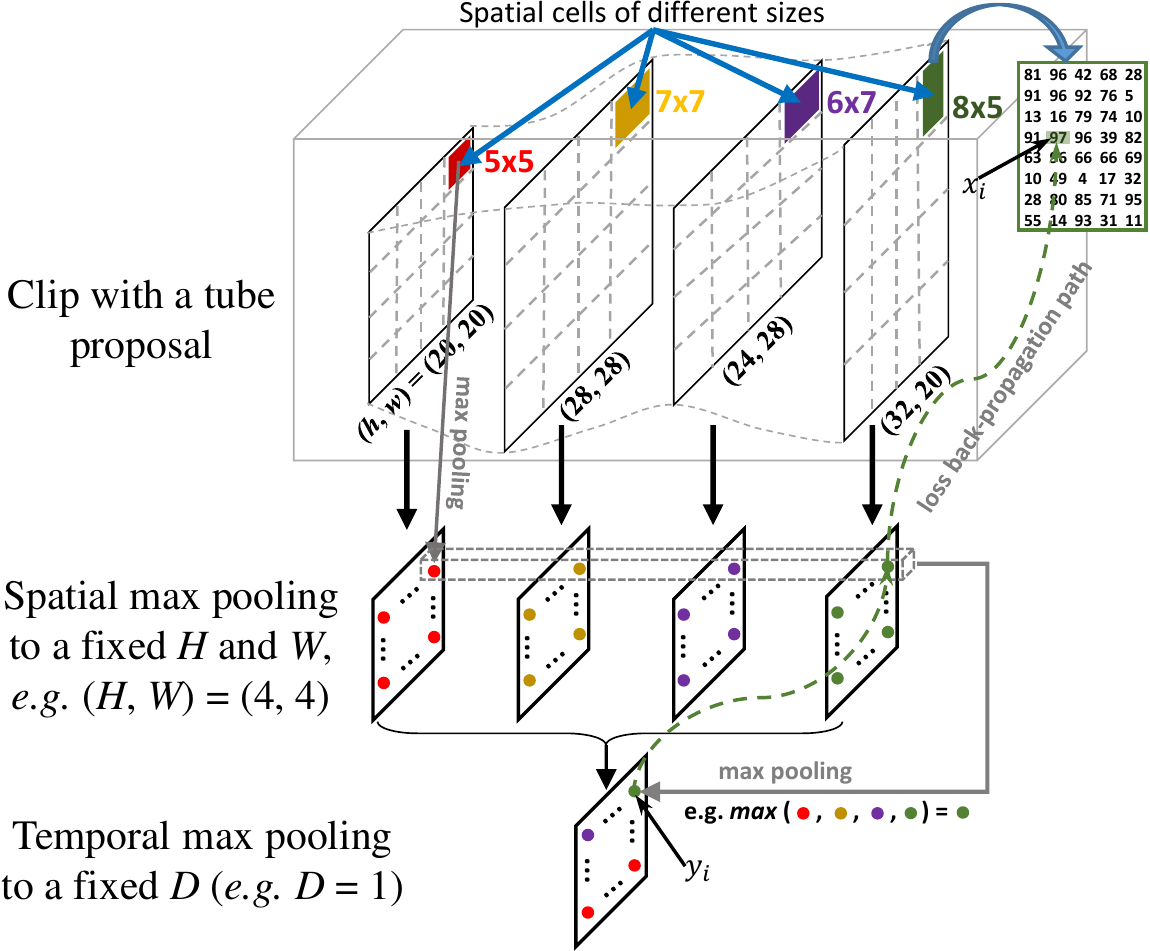}
\caption{Tube of interest pooling. In this example four feature maps of different sizes are first pooled spatially to $4\times4$ then pooled temporally to a fixed size of $1$. The cell size for spatial pooling in each feature map varies depending on the size of the feature map.  ``\textcolor{ao(english)}{$\dashrightarrow$}" indicates the loss back-propagation path of a ToI output, $y_i$ (\protect \tikz \protect \draw[ao(english),fill=ao(english)] (0,0) circle (.4ex);), to its corresponding input, $x_i$.  }
\label{fig:toi_pool}
\end{figure}

Since a video is processed clip by clip, action tube proposals with various spatial and temporal sizes are generated for different clips. These clip proposals need to be linked into a  sequence of tube proposal, which  is used for action label prediction and localization. To produce a fixed length feature vector, we propose a new pooling layer -- {\bf Tube-of-Interest} (ToI) pooling layer. The ToI pooling layer is a 3D generalization of Region-of-Interest (RoI) pooling layer of R-CNN. The classic max pooling layer defines the kernel size, stride and padding which determine the shape of the output. In contrast, for RoI pooling layer, the output shape is fixed first, then the kernel size and stride are determined accordingly. Compared to RoI pooling which takes 2D feature map and 2D regions as input, ToI pooling deals with feature cube and  3D tubes. Denote the size of a feature cube as $d\times h\times w$, where $d$, $h$ and $w$ respectively represent the depth, height and width of the feature cube. A ToI in the feature cube is defined by a $d$-by-$4$ matrix, which is composed of $d$ boxes distributed in $d$ feature maps. The boxes are defined by a four-tuple $(x_{1}^{i}, y_{1}^{i}, x_{2}^{i}, y_{2}^{i})$ that specifies the top-left and bottom-right corners in the $i$-th feature map. Since the $d$ bounding boxes may have different sizes, aspect ratios and positions, in order to apply spatio-temporal pooling, pooling in spatial and  temporal domains are performed separately. First, the $h\times w$ feature maps are divided  into $H\times W$ bins, where each bin corresponds to a cell with size of approximately $h/H\times w/W$. In each cell, max pooling is applied to select the maximum value. Second, the spatially pooled $d$ feature maps are temporally divided into $D$ bins. Similar to the first step, $d/D$ adjacent feature maps are grouped together to perform the standard temporal max pooling.  As a result the fixed output size of ToI pooling layer is $D \times H \times W$. A graphical illustration of ToI pooling is presented in Figure \ref{fig:toi_pool}.

Back-propagation of ToI pooling layer routes the derivatives from output back to the input as shown in Figure\ref{fig:toi_pool}. Assume $x_{i}$ is the $i$-th activation to the ToI pooling layer, and $y_{j}$ is the $j$-th output. Then the partial derivative of the loss function ($L$) with respect to each input variable $x_i$ can be expressed as:
\begin{align}
\frac{\partial L}{\partial x_{i}} = \frac{\partial L}{\partial y_{j}}\frac{\partial y_{j}}{\partial x_{i}} = \sum_{j}[i = f(j)]\frac{\partial L}{\partial y_{j}}.
\end{align}
Each pooling output $y_j$ has a corresponding input position $i$. We use a function $f(\cdot)$ to represent the \texttt{argmax} selection from the ToI. Thus, the gradient from the next layer $\partial L / \partial y_{j}$ is passed back to only that neuron which achieves the max $\partial L / \partial x_{i}$. 

\section{T-CNN for Action Detection}
\label{sec:approach}

\begin{figure}[!ht]
	\centering
	\includegraphics[width=0.9\linewidth]{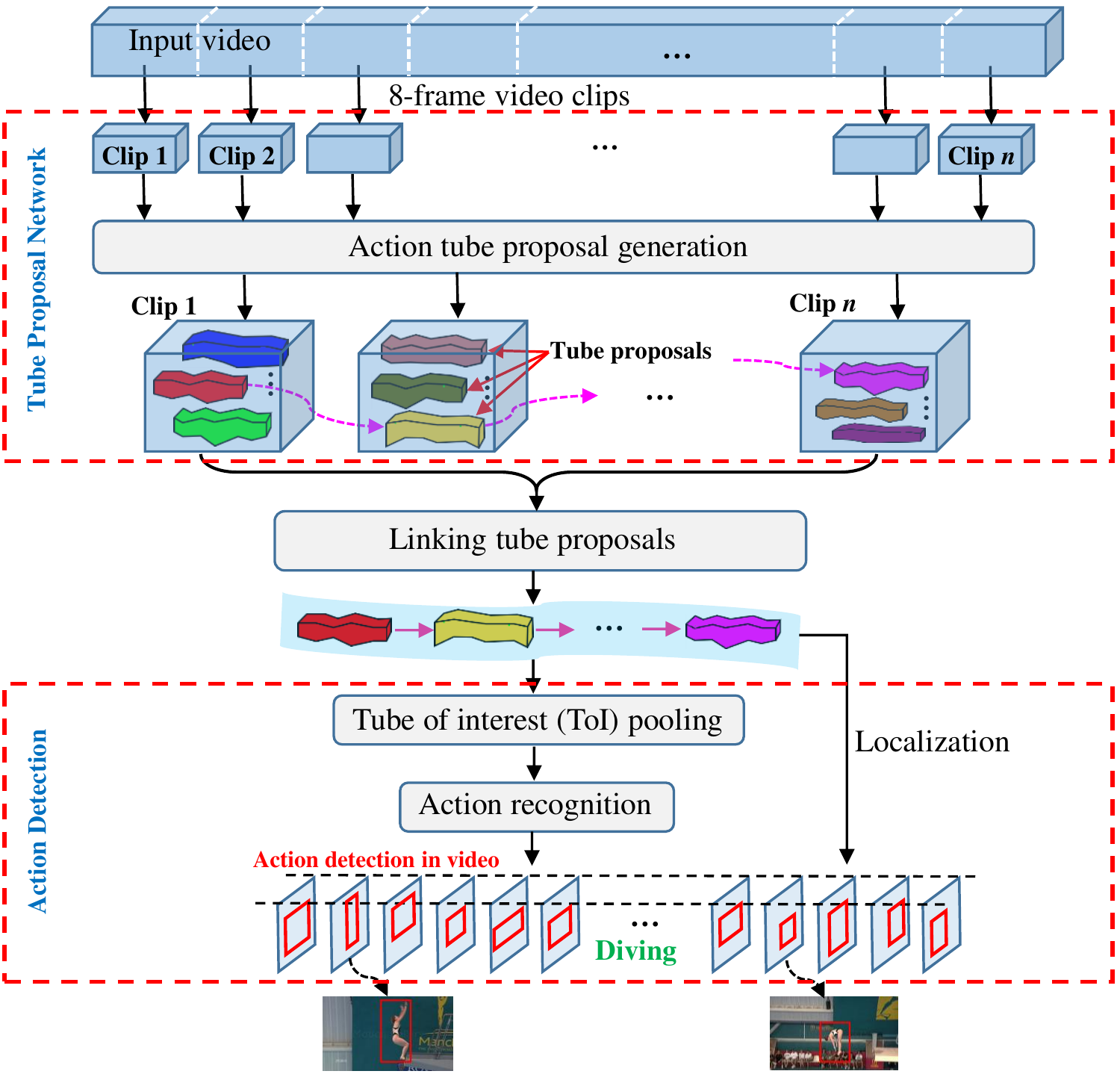}
	\caption{An overview of the proposed Tube Convolutional Neural Network (T-CNN) for action detection. First, an input video is divided into equal length clips of $8$ frame each and fed to Tube Proposal Network to generate tube proposals.  Next, these proposals are then linked into larger tubes covering full actions and fed to Action Detection network. Finally,   Action Detection network employs TOI pooling to recognize and localize the action.}
	\label{fig:Teaser}
\end{figure}

As shown in Figure \ref{fig:Teaser}, our T-CNN is an end-to-end deep learning framework that takes video clips as input. The core component is the Tube Proposal Network (TPN) (see Figure \ref{fig:TPN}) to produce tube proposals for each clip. Linked tube proposal sequences are used for  spatio-temporal action detection and action recognition in the video.


\subsection{Tube Proposal Network}
\label{sub-sec: tpn}
For a 8-frame video clip, 3D convolution and 3D pooling are used to extract spatio-temporal feature cube.
In 3D CNN, convolution and pooling are performed spatio-temporally. Therefore, the temporal information of the input video is preserved. Our 3D CNN consists of seven 3D convolution layers and four 3D max-pooling layers. We denote the kernel shape of 3D convolution/pooling by $d\times h\times w$, where $d, h, w$ are depth, height and width, respectively. In all convolution layers, the kernel sizes are $3\times 3\times 3$, padding and stride remain as $1$. The numbers of filters are 64, 128 and 256 respectively in the first $3$ convolution layers and $512$ in the remaining convolution layers.  The kernel size is set to $1\times 2\times 2$ for the first 3D max-pooling layer, and $2\times 2\times 2$ for the remaining 3D max-pooling layers. The details of network architecture are presented in Table \ref{tab:architecture}.  We use the C3D model \cite{c3d} as the pre-trained model and  finetune it on each dataset in our experiments.

After 3D conv5, the temporal size is reduced to 1 frame (\ie~feature cube with depth $D = 1$). \textit{In the feature cube, each frame/slice consists of a number of channels specified in Table \ref{tab:architecture}. Here, we drop the number of channels for ease of explanation.} Following Faster R-CNN, we generate bounding box proposals based on the conv5 feature cube\footnote{Since the depth of conv5 feature cube is reduced to 1, the ToI pooling after that essentially reduces to RoI pooing, a special case of ToI. For consistency, we use ToI throughout the entire framework.}.

\textbf{Anchor bounding boxes selection.} In Faster R-CNN, the bounding box dimensions are hand picked, \ie~9 anchor boxes with 3 scales and 3 aspect ratios. We can directly adopt the same anchor boxes in our T-CNN framework. However, it has been shown in \cite{yolo9000} that if we choose better bounding box priors for the network, it helps the network learn better to predict good detections.
Therefore, instead of choosing hand-picked anchor boxes, we apply k-means
clustering on  bounding boxes in the training set to learn $12$ anchor boxes (\ie~cluster centers). This data driven anchor box selection approach is adaptive to different datasets.

Each bounding box (bbx) is associated with an ``actionness" score, which measures the probability that  the  bbx corresponds to a valid action. We assign a binary class label (of being an action or not) to each bounding box. Bounding boxes with actionness scores smaller than a threshold are discarded. In the training phase, the bbx which has an Intersection-over-Union (IoU) overlap higher than 0.7 with any ground-truth  bbx, or has the highest IoU overlap with a ground-truth box (the later condition is considered in case the former condition may find no positive sample) is taken as a positive bounding box proposal.

\begin{figure*}[t]
	\centering
	\includegraphics[width=0.8\linewidth]{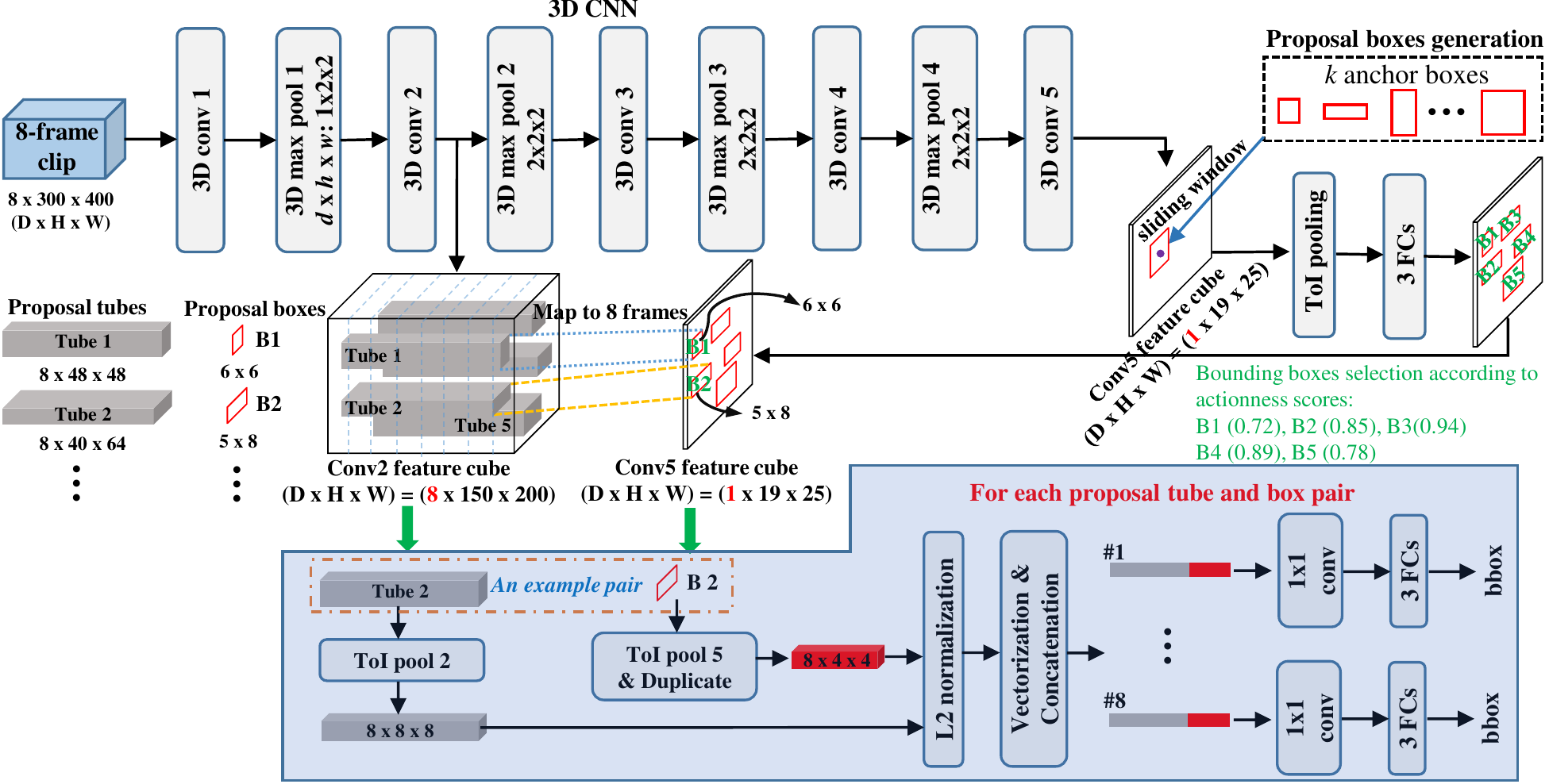}
	\caption{The tube proposal network (TPN) takes a 8-frame clip as input and applies 3D convolution and max-pooing to extract spatio-temporal features. conv5 feature cube are used to generate bounding box proposals. Those with high actionness scores are mapped to conv2 feature cube (contains 8 frames information) at the corresponding positions to get tube proposals. Each proposal tube and box pair are aggregated after separate ToI pooling. Then bounding box regression is performed for each frame. Note that the channel dimension is omitted in this figure (see Table~\ref{tab:architecture}).}
	\label{fig:TPN}
\end{figure*}

\textbf{Temporal skip pooling.} Bounding box proposals generated from conv5 feature tube can be used for frame-level action detection by bounding box regression. However, due to  temporal max pooling (8 frames to 1 frame), the temporal order  of the original 8 frames is lost. Therefore, we use temporal skip pooling to inject the temporal order  for frame-level detection. Specifically, we map each positive bounding box generated from conv5 feature cube into conv2 feature cube which has 8 feature frames/slices. Since these 8 feature slices correspond to the original 8 frames in a video clip, the temporal order  is preserved. As a result, if there are 5 bounding boxes in conv5 feature cube for example, 5 scaled bounding boxes are mapped to each conv2 feature slice at the corresponding locations. This creates 5 tube proposals as illustrated in Figure \ref{fig:TPN}, which are paired with the corresponding 5 bounding box proposals for frame-level action detection. To form a fixed feature maps, ToI pooling is applied to the variable size tube proposals as well as the bounding box proposals. Since a tube proposal covers 8 frames in Conv2, the ToI pooled bounding box from Conv5 is duplicated 8 times to form a tube. We then apply L2 normalization to the paired two tubes, and vectorize and concatenate them.
Since we use the C3D model \cite{c3d} as the pre-trained model, we apply a 1x1 convolution to match the input dimension of fc6.
Three fully-connected (FC) layers process each descriptor and produce the output: displacement of height, width and  2D center of each bounding box (``bbox") in each frame. The regression loss measures the differences between ground-truth and predicted bounding boxes,  represented by a 4D vector ($\Delta center_x$, $\Delta center_y$, $\Delta width$, $\Delta height$). The sum of them for all bounding boxes is the regression loss of the whole tube.
Finally, a set of refined tube proposals by adding the displacements of height, width and center are generated as an output from the TPN representing potential spatio-temporal action localization of the input video clip.

\subsection{Linking Tube Proposals}
\label{subsec: link}

\begin{figure}[!ht]
\centering
\includegraphics[width=0.95\columnwidth]{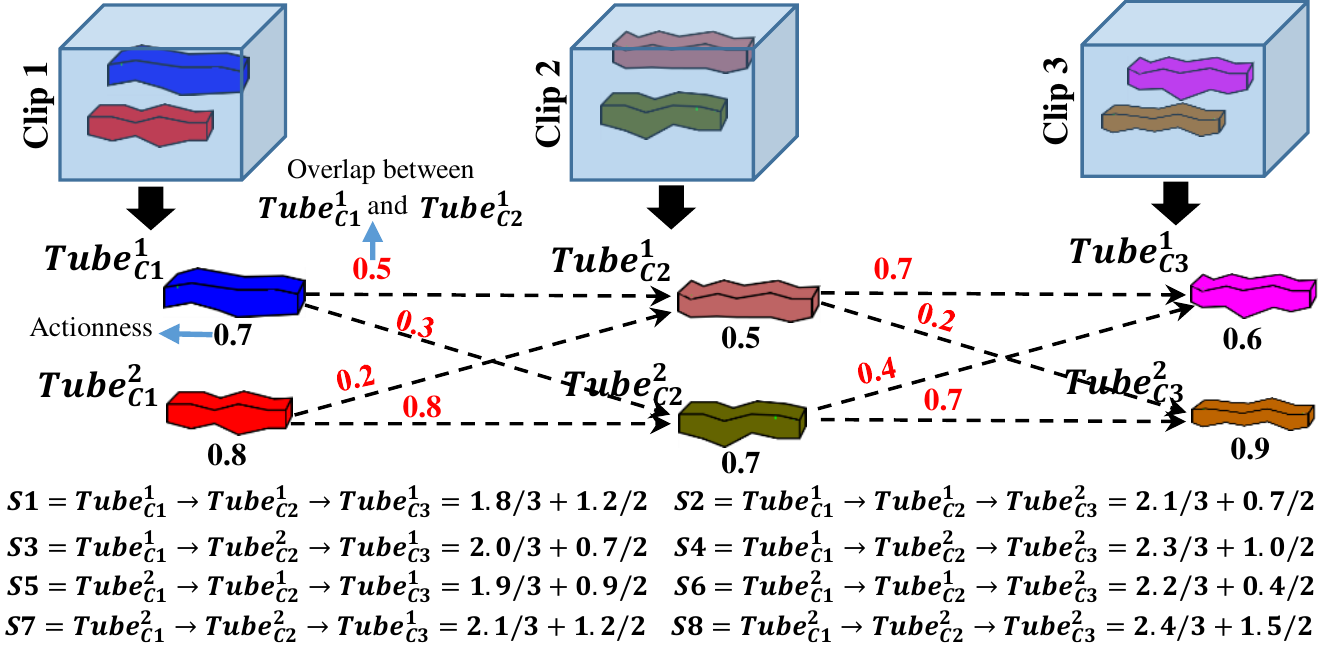}
\caption{An example of linking tube proposals in video clips using network flow. In this example, there are three video clips and each has two tube proposals, resulting in 8 video proposals. Each video proposal has a score, \eg~$S1, S2, ..., S8$, which is computed according to Eq. (2).}
\label{fig:link}
\end{figure}

A set of tube proposals are obtained for each video clip after the TPN. We then link these tube proposals to form a sequence of proposals for spatio-temporal action localization of the entire video.
Each tube proposal from different clips can be linked in a tube proposal sequence (\ie~video tube proposal) for action detection. However, not all combinations of tube proposals can correctly capture the complete action. For example, a tube proposal in one clip may contain the action and a tube proposal in the following clip may only capture the background.  Intuitively, the content within the selected tube proposals should capture an action and connected tube proposals in any two consecutive clips should have a large temporal overlap. Therefore, two criteria are considered when linking tube proposals: actionness and overlap scores. Each video proposal is then assigned a score defined as follows:
\begin{equation}
S=\frac{1}{m}\sum_{i=1}^m Actionness_i+\frac{1}{m-1}\sum_{j=1}^{m-1} Overlap_{j,j+1},
\end{equation}
where $Actionness_i$ denotes the actionness score of the tube proposal from the $i$-th clip, $Overlap_{j,j+1}$ measures the overlap between the linked two proposals respectively from the $j$-th and $(j+1)$-th clips, and $m$ is the total number of video clips. As shown in Figure \ref{fig:TPN}, each bounding box proposal from conv5 feature cube is associated with an actionness score. These actionness scores are inherited by the corresponding tube proposals from conv2 feature cube. The overlap between two tube proposals is calculated based on the IoU (Intersection Over Union) of the last frame of the $j$-th tube proposal and the first frame of the $(j+1)$-th tube proposal. The first term of $S$ computes the average actionness score of all tube proposals in a video proposal and the second term computes the average overlap between the tube proposals in every two consecutive video clips. Therefore, we ensure the linked tube proposals can encapsulate the action and at the same time have temporal consistency. An example of linking tube proposals and computing scores is illustrated in Figure \ref{fig:link}.
We choose a number of linked proposal sequences with highest scores in a video (see more details in Sec.~\ref{subsec:basic_settings}).

\subsection{Action Detection}
\label{subsec: recognition}
After linking tube proposals, we get a set of linked sequences of  tube proposal, which represent potential action instances. The next step is to classify these linked tube proposal sequences. The tube proposals in the linked sequences may have different sizes. In order to extract a fixed length feature vector from each of the linked proposal sequence, our proposed ToI pooling is utilized. Then the ToI pooling layer is followed by two FC layers and a drop-out layer. The dimension of the last FC layer is $N+1$ ($N$ action classes and $1$ background class). 

\section{Bottom-up Action Detection in Videos}
 \label{sec:seg_overview}
\begin{figure*}[!t]
\centering
\includegraphics[width=0.95\linewidth]{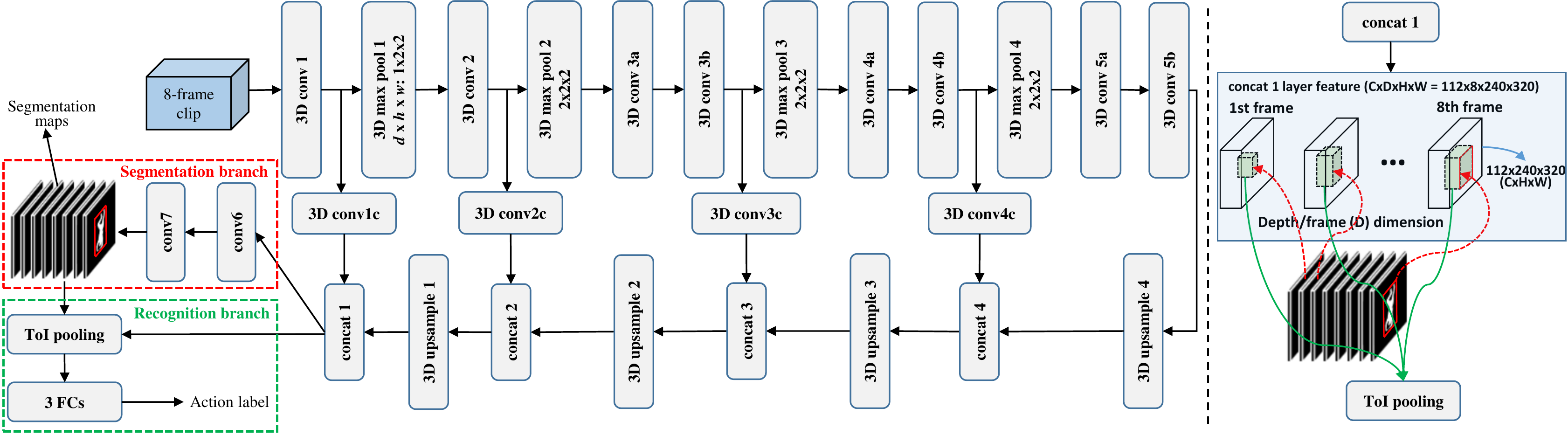}
\caption{The framework of ST-CNN for action detection. An encoder-decoder 3D CNN architecture is employed. The segmentation branch produces a binary segmentation map (action foreground vs. background) for each frame of the input video clip. The foreground pixels are used to infer frame-wise bounding boxes (action localization), based on which the feature tube is extracted from ``concat1" feature cube as an input to the recognition branch (see the right part of this figure for details).   }
\label{fig:segmentation}
\end{figure*}

The proposed T-CNN approach is able to detect actions in videos and put bounding boxes for localization in each frame. As discussed before, T-CNN as a top-down approach relies on exhaustive search in the whole frame and appropriate bounding boxes selection. It has been shown that bottom-up approaches  which directly operate on group of pixels \eg~through super-voxel or super pixel segmentation are more efficient for action detection. Also, it is obvious that pixel-wise action segmentation maps provide finer human silhouettes than bounding boxes, since bounding may also include background pixels.

To achieve this goal, we develop a ST-CNN (Segmentation Tube CNN) approach to automatically localize and segment the silhouette of an actor for action detection.
Figure~\ref{fig:segmentation} shows the network structure of the proposed ST-CNN. It is an end-to-end 3D CNN, which builds upon an encoder-decoder structure like the SegNet \cite{badrinarayanan2015segnet} for image semantic segmentation. A video is divided into 8-frame clips as input to the network. On the encoder side, 3D convolution and max pooling are performed. Due to 3D max pooling, the spatial and temporal sizes are reduced. In order to generate the pixel-wise segmentation map for each frame in the original size, 3D upsampling is used in the decoder to increase the resolution of feature maps. To capture spatial and temporal information at different scales, a concatenation with the corresponding feature maps from the encoder is employed after each 3D upsampling layer. Finally, a segmentation branch is used for pixel-wise prediction (\ie~background or action foreground) for each frame in a clip.
Recognition branch takes the segmentation maps (output of the segmentation branch), where the foreground segmentation maps (action regions) are converted into bounding boxes, and the feature cube of the last concatenation layer ``concat1", to extract the feature tube of the action volume. ToI pooling is applied to the feature tube and followed by three FC layers for action recognition. The detailed network configuration is presented in Table~\ref{tab:mtcnn_architecture}.

\begin{table}
\caption{Network architecture of ST-CNN. Note that conv6 and conv7 (in the segmentation branch) are applied to each frame of the concat1 layer feature cube ($112 \times 8 \times 240 \times 320$) to produce 8 segmentation maps.}
\begin{center}
\begin{tabular}{lcc}
\hline
name            & kernel dims        & output dims \\
 &  ($d\times h \times w$) & ($C\times D\times H\times W$) \\
\hline
conv1           & $3\times3\times3$ & $64\times8\times240\times320$ \\
max-pool1       & $1\times2\times2$ & $64\times8\times120\times160$ \\
conv2           & $3\times3\times3$ & $128\times8\times120\times160$ \\
max-pool2       & $2\times2\times2$ & $128\times4\times60\times80$ \\
conv3a          & $3\times3\times3$ & $256\times4\times60\times80$ \\
conv3b          & $3\times3\times3$ & $256\times4\times60\times80$ \\
max-pool3       & $2\times2\times2$ & $256\times2\times30\times40$ \\
conv4a          & $3\times3\times3$ & $512\times2\times30\times40$ \\
conv4b          & $3\times3\times3$ & $512\times2\times30\times40$ \\
max-pool4       & $2\times2\times2$ & $512\times1\times15\times20$ \\
conv5a          & $3\times3\times3$ & $512\times1\times15\times20$ \\
conv5b          & $3\times3\times3$ & $512\times1\times15\times20$ \\
\hline
upsample4       & $3\times3\times3$ & $64\times2\times30\times40$ \\
conv4c          & $3\times3\times3$ & $448\times2\times30\times40$ \\
upsample3       & $3\times3\times3$ & $64\times4\times60\times80$ \\
conv3c          & $3\times3\times3$ & $448\times4\times60\times80$ \\
upsample2       & $3\times3\times3$ & $64\times8\times120\times160$ \\
conv2c          & $3\times3\times3$ & $128\times8\times120\times160$ \\
upsample1       & $3\times3\times3$ & $48\times8\times240\times320$ \\
conv1c          & $3\times3\times3$ & $64\times8\times240\times320$ \\
\hline
conv6            & $1\times1$        & $4096\times 8 \times 240\times320$ \\
conv7            & $1\times1$        & $2\times 8 \times 240\times320$ \\
\hline
toi-pool        & --                & $112\times8\times8\times8$ \\
fc6             & --                & $4096$ \\
fc7             & --                & $4096$ \\
\hline
\end{tabular}
\end{center}
\label{tab:mtcnn_architecture}
\end{table}

Compared to T-CNN, ST-CNN is a bottom-up approach and has the following advantages for action detection.

$\bullet$ ST-CNN provides pixel-level localization for actions (\ie~action segmentation maps), which are more accurate than bounding boxes for action localization. Moreover, bounding boxes can be easily derived from the segmentation maps.

$\bullet$ ST-CNN avoids bounding box regression used in T-CNN, which is hard to converge. Specifically, to learn a sufficient regression model, large amount of data is required. 
In contrast,  less data is required for binary (foreground vs. background) classification and convergence is easier to reach. In our case, lack of annotated training video data causes difficulty in training the bounding boxes regression model. Since ST-CNN uses  a classification model, which predicts foreground/background label for each pixel, the number of training samples is drastically increased, since one detection bounding box or action segmentation map includes hundreds of pixels.


\textbf{Connections between T-CNN and ST-CNN.} Although we do not employ TPN om ST-CNN  as in T-CNN, both frameworks explicitly use 3D convolutions as building blocks to learn spatio-temporal features  for action detection. The encoder part of ST-CNN shares the same network architecture of T-CNN for feature extraction. Moreover, ToI pooling is employed in both frameworks to handle variable bounding box sizes in tubes.

\subsection{3D Upsampling Layer}
\label{subsec: 3deconv}

\begin{figure}
\centering
\includegraphics[width=1.0\columnwidth]{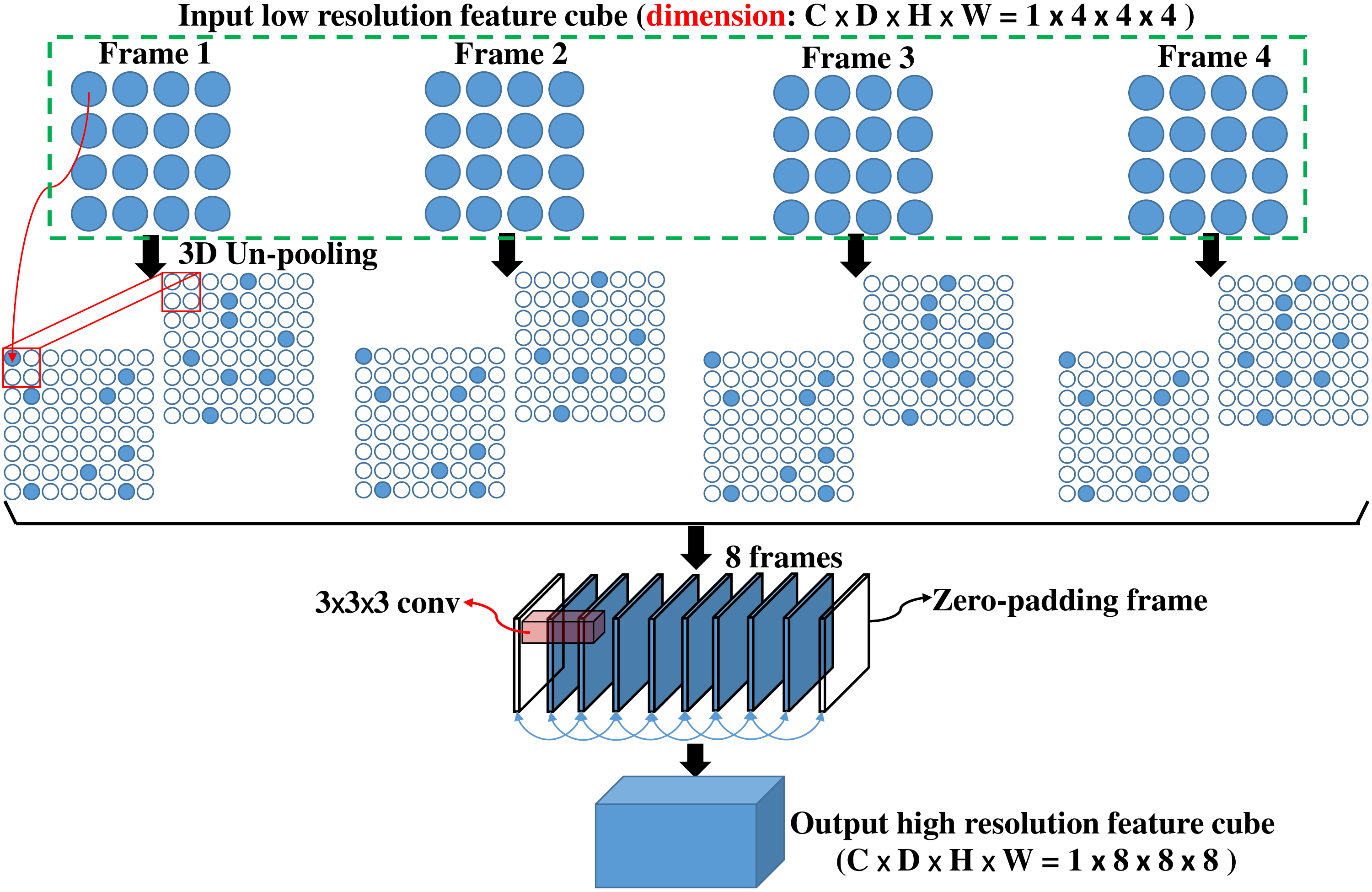}
\caption{The 3D ``un-pooling + convolution" operation with a upscale factor of 2 in both spatial and temporal dimensions. In this example, the input is a low resolution (LR) input feature cube ($C \times D \times H \times W = 1 \times 4 \times 4 \times 4 $). The un-pooling operation is used to generate two upscaled frames ($H \times W = 8 \times 8$) for each input frame. (Note that the un-pooling locations are randomly selected for demonstration purpose.) Then 3D convolution is applied to the resulting 8 frames and 2 zero-padding frames to produce the high resolution (HR) feature cube ($C \times D \times H \times W = 1 \times 8 \times 8 \times 8 $).  }
\label{fig:unpooling}
\end{figure}

A common method of recovering higher image resolution from max-pooling is to use un-pooling \cite{zeiler2014visualizing} followed by a convolution layer. The locations of the maximum values within max-pooling regions are recorded. The un-pooling layer places the values from the layer above to appropriate positions, according to the recorded max-pooling locations, as an approximation of reverse max-pooling. Deconvolution \cite{noh2015learning} is another approach to increase the resolution feature map, and has been explored in various tasks such as semantic segmentation \cite{noh2015learning}, optical flow estimation \cite{fischer2015flownet}, etc. A graphical illustration of the 3D un-pooling followed by 3D convolution is presented in Figure~\ref{fig:unpooling}.


\begin{figure}
\centering
\includegraphics[width=1.0\columnwidth]{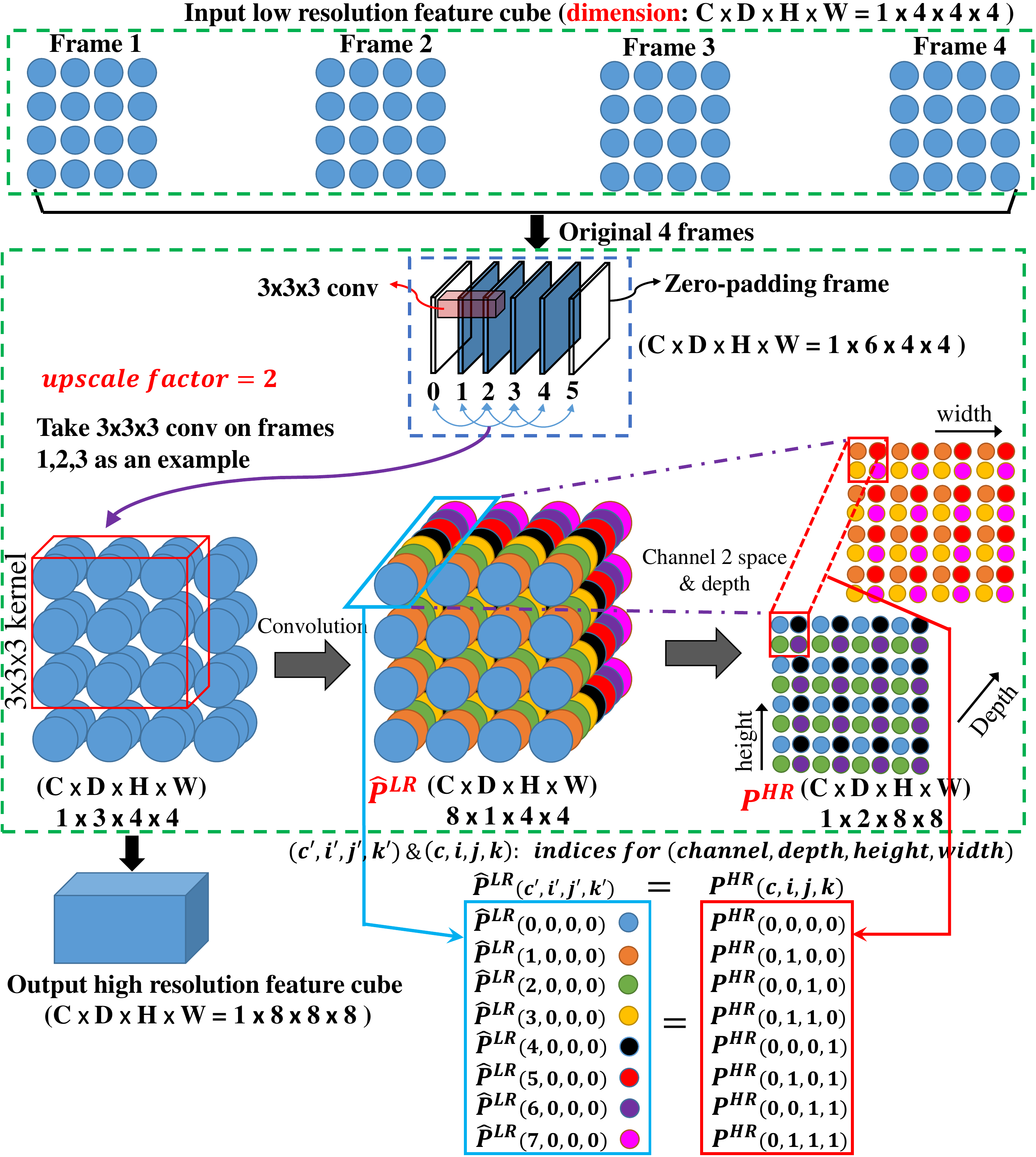}
\caption{The 3D sub-pixel convolution approach for feature cube upscaling. In this example, given the LR input feature cube with dimension ($C \times D \times H \times W = 1 \times 4 \times 4 \times 4 $). It is first added with two zero-padding frames. Then the $3 \times 3 \times 3$ convolution operation is applied to those frames (\ie~0-2, 1-3, 2-4, and 3-5). Take the 3D convolution on frames 1-3 as an example. We use 8 3D convolution kernels to generate $\hat{P}^LR$ with 8 channels.
Then channel to space \& depth reshape is performed, following Eqs.~\ref{eq:up} and \ref{eq:up-index}, to produce $P^{HR}$, where both the spatial dimension ($H \times W$) and the temporal dimension ($D$) are increased by a factor of 2. Therefore, the operations (3D convolution and channel-to-space\&depth reshape) on the input frames (0-2, 1-3, 2-4, and 3-5) result in an upscaled feature cube ($C \times D \times H \times W = 1 \times 8 \times 8 \times 8 $).}
\label{fig:up_illustration}
\end{figure}

Compared to un-pooling and deconvolution, sub-pixel convolution proposed in \cite{shi2016real} shows better performance for image super-resolution and is more computationally efficient. Therefore,
we adopt the sub-pixel convolution approach \cite{shi2016real} and generalize  it to 3D. An example of the 3D sub-pixel convolution operation to upscale an input low resolution feature cube by a factor of 2 in both spatial and temporal dimensions is demonstrated in Figure~\ref{fig:up_illustration}.

Formally, suppose we want to upsample the LR feature maps (or cube) by $p_d$, $p_h$ and $p_w$ times in depth (frame), height and width, respectively.
A convolution layer is first applied on the LR feature maps, $P^{LR}$ with dimension $(C \times D \times H \times W = C^L \times D^L \times H^L \times W^L)$, to expand the channels from $C^{L}$ to $C^H=p_{d}\times p_{h}\times p_{w}\times C^{L}$. In the example of Figure~\ref{fig:up_illustration}, $C^L=1$ and the upscale factor is 2, \ie~$p_d = 2$, $p_h = 2$ and $p_w = 2$. We denote the channel expanded feature maps as $\hat{P}^{LR}$. The spatial ($H \times W$) and temporal ($D$) resolutions of $\hat{P}^{LR}$ are the same as those of $P^{LR}$.
Second, subsequent channel-to-space\&depth transpose layer is placed on top of it to upsample the spatial and temporal dimensions by ($p_h$, $p_w$) and $p_d$ times, respectively. Let $P_{c,i,j,k}^{HR} (C \times D \times H \times W = C^H \times D^H \times H^H \times W^H)$ be the pixel value in the HR feature maps, located at $\left(c,i,j,k\right)$ denoting the position (channel, depth/frame, height, width). The pixels in the HR feature maps $P^{HR}$ are mapped from $\hat{P}^{LR}$ according to:
\begin{equation}\label{eq:up}
P^{HR}_{c,i,j,k} = \hat{P}^{LR}_{c',i', j', k'},
\end{equation}
where $c \in \{0,..., C^H-1\}$, $i \in \{0,..., D^H-1\}$, $j \in \{0,..., H^H-1\}$, and $k \in \{0,..., W^H-1\}$. Indices $c'$, $i'$, $j'$ and $k'$ for $\hat{P}^{LR}$ are defined as follows:
\begin{equation}\label{eq:up-index}
\left\{\begin{matrix}
c' = c \cdot p_{d}\cdot p_{h}\cdot p_{w} + \bmod(i,p_{d})
                    + p_{w}\cdot\bmod(j,p_{h}) \\
                   \hspace{-1.1in} + p_{w}\cdot p_{h}\cdot\bmod(k,p_{w})
\\ \hspace{-2.25in} i' = \lfloor i/p_{d}\rfloor
\\ \hspace{-2.24in} j' = \lfloor j/p_{h}\rfloor
\\ \hspace{-2.22in} k' = \lfloor k/p_{w}\rfloor
\end{matrix}\right.
\end{equation}

Note that 3D convolution is performed in the LR feature space in the sub-pixel convolution approach, whereas it is applied on the un-pooled sparse feature cube (in HR space) in the ``un-pooling + convolution" strategy. Therefore, the 3D sub-pixel convolution method is able to integrate more information in 3D convolution for feature cube upscaling.

\section{Experiments}
\label{sec:experiments}

To verify the effectiveness of the proposed 3D end-to-end deep learning framework for action detection and segmentation, we evaluate our approach on three trimmed video action detection datasets including UCF-Sports \cite{rodriguez2008action},
J-HMDB \cite{Jhuang:ICCV:2013}, UCF-101 \cite{THUMOS13}; one un-trimmed video action detection dataset -- THUMOS'14 \cite{THUMOS14}; and a video object segmentation dataset -- DAVIS'16 \cite{Perazzi2016}. In the following, we provide the implementation details of our networks, datasets and experiment settings, as well as show experimental comparisons of our method to the state-of-the-arts.

\subsection{Implementation Details}
\label{subsec:basic_settings}
\subsubsection{T-CNN for action detection}
The TPN and recognition network share weights in their common layers. Due to memory limitation, in the training phase, each video is divided into overlapping $8$-frame clips with resolution $300\times400$ and temporal stride $1$. When training the TPN, each anchor box is assigned a binary label. An anchor box which has the highest IoU overlap with a ground-truth box, or an anchor box that has an IoU overlap higher than 0.7 with any ground-truth box is assigned a positive label, and others are assigned negative labels. In each iteration, $4$ clips are fed into the network. Since the number of background boxes (\ie~negative boxes) is much larger than that of action boxes, we randomly select some of the negative boxes to balance the number of positive and negative samples in a batch. For recognition network training, we choose 40 linked proposal sequences with highest scores in a video as Tubes of Interest.

T-CNN model is trained in an alternative manner. First, we \textbf {initialize the TPN} based on the pre-trained model in \cite{c3d}, and use the generated proposals to \textbf{initialize the recognition network}. Next, the weights tuned by the recognition network are used to \textbf{update the TPN}. Finally, the tuned weights and proposals from the TPN are used for \textbf{finalizing the recognition network}. For UCF-Sports and J-HMDB, the learning rate of the networks is initialized as $10^{-3}$ and decreased to $10^{-4}$ after $30$k batches. Training terminates after $50$k batches. For UCF-101 and THUMOS'14, the learning rate is initialized as $10^{-3}$ and decreased to $10^{-4}$ after $60$k batches. Training terminates after $100$k batches.

During testing, each video is divided into non-overlapping $8$-frame clips. If the number of frames in the video cannot be divided by $8$, we pad zeros after the last frame to make it divisible. $40$ tube proposals with highest actionness confidence through TPN are chosen for the linking process. Non-maximum suppression (NMS) is applied to the linked proposals to get the final action detection results.

\subsubsection{ST-CNN for action detection}
We use the same resolution $300\times400$ of the input videos without cropping and resizing. The detection task uses both losses (segmentation loss and recognition loss) as shown in Figure~\ref{fig:segmentation}. In the training phase, similar to T-CNN, videos are divided into overlapping $8$-frame clips with temporal stride $1$. For the J-HMDB dataset, the ground-truth annotations provide pixel-wise binary labels, where foreground is considered as positive and background as negative. These pixel-wise annotations are used for minimizing the segmentation loss. Once the segmentation is generated for each frame, a bounding box is obtained by including positive pixels (\ie~foreground) in each frame. Bounding boxes in each video clip form tubes of interest. Through ToI max pooling, the feature cube size is fixed for action recognition. The learning rate is initialized at $10^{-4}$ and decreased to $10^{-5}$ after $498,000$ batches where each batch contains $4$ clips. Training terminates after $500,000$ batches. During testing, the incoming video is divided into non-overlapping $8$-frame clips first. Then, the segmentation branch predict the foreground/background for all the pixels (\textbf{action segmentation -- localization}). Finally, we take the boundary of predicted foreground pixels in each frame to form the Tube-of-Interest, which is fed into the ToI pooling layer to predict the action label (\textbf{action recognition}).

\subsubsection{ST-CNN for video object segmentation}
Since ST-CNN already includes a segmentation loss, it can be easily applied to the video object segmentation task, \ie~segmenting the primary foreground object in each video frame. The benchmark dataset, DAVIS'16 \cite{Perazzi2016}, is employed for evaluation. In this experiment, only segmentation loss is considered as shown in Figure~\ref{fig:segmentation} since action recognition is not involved. Similar to the detection task, each batch contains $4$ clips, but the learning rate is initialized at $10^{-3}$ and decreased to $10^{-4}$ after $100,000$ batches. The training process terminates after $120,000$ batches.

\subsection{Action Detection Experimental Results}
\label{subsec:results}

\subsubsection{UCF-Sports}
\label{subsubsec:ucfsports}
UCF-Sports dataset contains 150 short videos of 10 different sport classes. Videos are trimmed and the action and bounding boxes annotations are provided for all frames. We follow the standard training and test split defined in \cite{lan2011discriminative} to carry out the evaluation.

\begin{figure*}[!thb]
\centering
\includegraphics[width=0.95\linewidth]{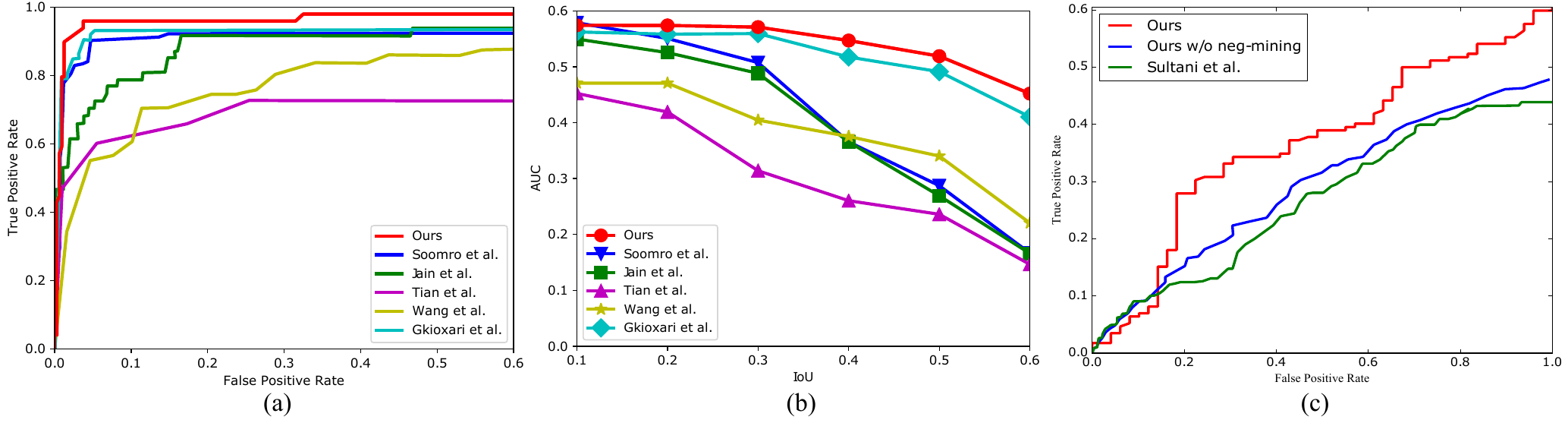}
\caption{The ROC and AUC curves for UCF-Sports dataset \cite{rodriguez2008action} are shown in (a) and (b), respectively. The results are shown for Jain \etal \cite{jain2014action} (green), Tian \etal \cite{tian2013spatiotemporal} (purple), Soomro \etal \cite{soomro2015action} (blue), Wang \etal \cite{wang2014video} (yellow), Gkioxari \etal \cite{gkioxari2015finding} (cyan) and Proposed Method (red). (c) shows the mean ROC curves for four actions of THUMOS'14. The results are shown for Sultani \etal \cite{Sultani_2016_CVPR} (green), the proposed T-CNN (red) and T-CNN without negative mining (blue).}
\label{fig:curve_result}
\end{figure*}

We use the usual IoU criterion and generate ROC curve in Figure \ref{fig:curve_result}(a) when overlap equals to $\alpha = 0.2$. Figure \ref{fig:curve_result}(b) illustrates AUC (Area-Under-Curve) measured with different overlap criterion. In direct comparison, our T-CNN clearly outperforms all the competing methods shown in the plot. We are unable to directly compare the detection accuracy against Peng \etal \cite{peng2016multi} in the plot, since they do not provide the ROC and AUC curves. As shown in Table \ref{tab:per-class}, the frame level mAP of our approach outperforms theirs in 8 actions out of 10. Moreover, by using the same metric, the video mAP of our approach reaches 95.2 ($\alpha = 0.2$ and $0.5$), while they report 94.8 ($\alpha = 0.2$) and 94.7 ($\alpha = 0.5$).

\begin{table*}[!thb]
\caption{mAP for each class of UCF-Sports. The IoU threshold $\alpha$ for frame m-AP is fixed to $0.5$.}
\begin{center}
\small
\begin{tabular}{lccccccccccc}
\hline
                                                    & Diving    & Golf  & Kicking   & Lifting   & Riding    & Run   & SkateB.   & Swing     & SwingB.   & Walk  & mAP \\
\hline
Gkioxari \etal \cite{gkioxari2015finding}           & 75.8      & 69.3  & 54.6      & 99.1      & 89.6      & 54.9  & 29.8      & 88.7      & 74.5      & 44.7  & 68.1 \\
Weinzaepfel \etal \cite{weinzaepfel2015learning}    & 60.71     & 77.55 & 65.26     & {\bf 100.00}& 99.53   & 52.60 & 47.14     & 88.88     & 62.86     & 64.44 & 71.9 \\
Peng \etal \cite{peng2016multi}                     & {\bf 96.12}& 80.47& 73.78     & 99.17     & 97.56     & 82.37 & 57.43     & 83.64     & 98.54     & 75.99 & 84.51 \\
Kalogeiton \etal \cite{kalogeiton2017action}    & --    & --    & --    & --    & --    & --    & --    & --    & --    & --    & {\bf 87.7} \\
Ours (T-CNN)                                        & 84.38     &  {\bf 90.79}& 86.48& 99.77    & {\bf 100.00}& {\bf 83.65} & {\bf 68.72}& 65.75     & {\bf 99.62}     & {\bf 87.79} & 86.7 \\
Ours (ST-CNN*)                 & 70.9      & 89.1  & {\bf 90.7}& 89.7      & 99.6      & 71.1  & 80.4      & {\bf 89.3}& 86.7      & 77.5  & 84.5 \\
\hline
\end{tabular}
\end{center}

\label{tab:per-class}
\end{table*}

\subsubsection{J-HMDB}
J-HMDB consists of 928 videos with 21 different actions. All the video clips are well trimmed. There are three train-test splits and the evaluation is done on the average results over the three splits. The experimental results comparison is shown in Table \ref{tab:jhmdb}. We report our results using 3 metrics: frame-mAP, the average precision of detection at frame level as in \cite{gkioxari2015finding}; video-mAP, the average precision at video level as in \cite{gkioxari2015finding} with IoU threshold $\alpha=0.2$ and $\alpha=0.5$. It is evident that our T-CNN consistently outperforms the state-of-the-art approaches in terms of all three evaluation metrics.

\begin{table}[!ht]
\caption{Comparison of the state-of-the-art approaches on J-HMDB. The IoU threshold $\alpha$ for frame m-AP is fixed to 0.5.}
\begin{center}
\small
\begin{tabular}{lccc}
\hline
                                                    & f.-mAP        & v.-mAP        & v.-mAP  \\
                                                    & ($\alpha=0.5$)& ($\alpha=0.2$)& ($\alpha=0.5$) \\
\hline
Gkioxari \etal \cite{gkioxari2015finding}           & 36.2          & --            & 53.3 \\
Weinzaepfel \etal \cite{weinzaepfel2015learning}    & 45.8          & 63.1          & 60.7 \\
Peng \etal \cite{peng2016multi}                     & 58.5          & 74.3          & 73.1 \\
Kalogeiton \etal \cite{kalogeiton2017action}        & {\bf 65.7}    & 74.2      & 73.7 \\
Singh \etal \cite{singh2016online}        & --    & 73.8      & 72.0 \\
Ours (T-CNN)                                        & 61.3          & 78.4   & 76.9 \\
Ours (ST-CNN)                                   & 64.9          & {\bf 78.6 }   & {\bf 78.3} \\
Ours (un-pool)                                       & 57.1          & 71.6            & 73.9 \\
\hline
\end{tabular}
\end{center}
\label{tab:jhmdb}
\end{table}

\subsubsection{THUMOS'13 (UCF 101)}
\label{subsubsec:ucf101}
UCF-101 dataset with 101 actions is commonly used for action recognition. For action detection task, a subset (THUMOS'13) of $24$ action classes and $3,207$ videos have spatio-temporal annotations \cite{singh2016online}. Similar to other methods, we perform the experiments on the first train/test split only. We report our results in Table \ref{tab:ucf101} with 3 metrics: frame-mAP, video-mAP ($\alpha=0.2$) and video-mAP ($\alpha=0.5$). Our approach again yields the best performance. Moreover, we also report the action recognition results of T-CNN on the above three datasets in Table \ref{tab:rec}.

\begin{table}[!ht]
\caption{Comparison of the state-of-the-art on UCF-101 (24 actions). The IoU threshold $\alpha$ for frame m-AP is fixed to 0.5.}
\begin{center}
\footnotesize
\begin{tabular}{lccccc}
\hline
                                        & f.-mAP & \multicolumn{4}{c}{video-mAP} \\
IoU th.                                 &           & $0.05$   & $0.1$    & $0.2$    & $0.3$ \\
\hline
Weinzaepfel \etal \cite{weinzaepfel2015learning}    & 35.84     & 54.3    & 51.7 & 46.8      & 37.8 \\
Peng \etal \cite{peng2016multi}                     & 65.73     & {\bf 78.76}& 77.31             & 72.86     & 65.7 \\
Kalogeiton \etal \cite{kalogeiton2017action}        & 67.1        & --      & -- & {\bf77.2}    & -- \\
Singh \etal \cite{singh2016online}        & --        & --      & -- & {73.5}    & -- \\
Ours                                                & {\bf 67.3}& 78.2    & {\bf 77.9}       & 73.1 & {\bf 69.4} \\
\hline
\end{tabular}
\end{center}
\label{tab:ucf101}
\end{table}


\subsubsection{THUMOS'14}
To further validate the effectiveness of our T-CNN approach for action detection, we evaluate it using the untrimmed videos from the THUMOS'14 dataset \cite{THUMOS14}. The THUMOS'14 spatio-temporal localization task consists of $4$ classes of actions: BaseballPitch, golfSwing, TennisSwing and ThrowDiscus. There are about $20$ videos per class and each video contains $500$ to $3,000$ frames. The videos are divided into validation set and test set, but only videos in the test set have spatial annotations provided by \cite{Sultani_2016_CVPR}. Therefore, we use samples corresponding to those 4 actions in UCF-101 with spatial annotations to train our model.

In untrimmed videos, there often exist other unrelated actions besides the action of interests. For example, ``walking" and ``picking up a golf ball" are considered as unrelated actions when detecting ``GolfSwing" action. We denote clips which have positive ground truth annotation as positive clips, and the other clips as negative clips (\ie~clips containing unrelated actions). If we randomly select negative samples for training, the number of boxes on unrelated actions is much smaller than that for background boxes (\ie~boxes capturing only background). Thus the trained model will have no capability to distinguish action of interest and unrelated actions.

To this end, we introduce a so called negative sample mining process. Specifically, when initializing the TPN, we only use positive clips. Then we apply the model on the whole training video (consisting of both positive and negative clips). Most false positives in negative clips should include unrelated actions to help our model learn the difference between action of interest and unrelated actions. Therefore we select boxes in negative clips with highest scores as \textbf{hard negatives}, since low scores probably related to the image background and those background clips do not contribute to the model to distinguish between action of interest and unrelated actions. In the TPN updating procedure, we choose 32 boxes which have IoU with any ground truth greater than 0.7 as positive samples and randomly pick another 16 samples as negative. We also select 16 samples from the hard negative pool as negatives. Therefore, we efficiently train a model, which is able to distinguish not only action of interest from background, but also action of interest from unrelated actions.

\begin{figure*}[!t]
\centering
\includegraphics[width=0.95\linewidth]{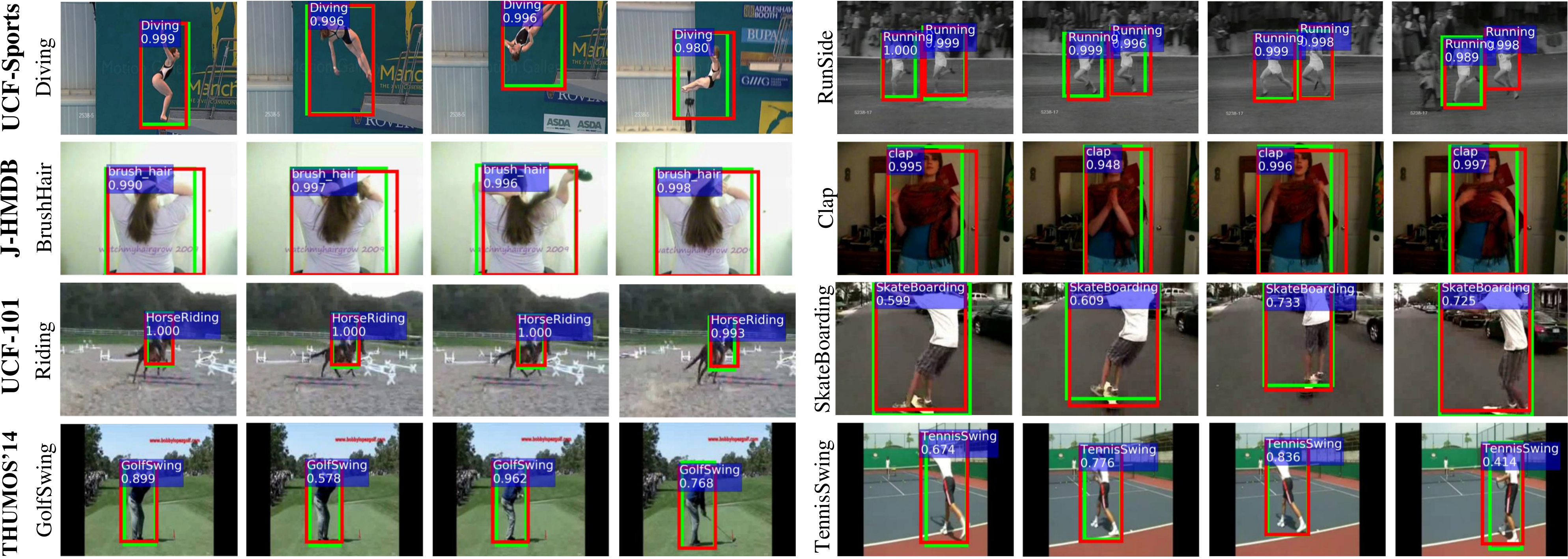}
\caption{Action detection results obtained by T-CNN on UCF-Sports, JHMDB, UCF-101 and THUMOS'14. Red boxes show the detections in the corresponding frames, and green boxes show ground truth. The predicted labels are overlaid. }
\label{fig:results}
\end{figure*}

The mean ROC curves of different methods for THUMOS'14 action detection are plotted in Figure \ref{fig:curve_result}(c). Our method without negative mining performs better than the baseline method Sultani \etal \cite{Sultani_2016_CVPR}. Additionally, with negative mining, the performance is further boosted. For qualitative results, we shows examples of detected action tubes in videos from UCF-Sports, JHMDB, UCF-101 (24 actions) and THUMOS'14 datasets (see Figure \ref{fig:results}). Each block corresponds to a different video that is selected from the test set. We show the highest scoring action tube for each video.


\begin{table}[!tb]
\caption{Action recognition results of our T-CNN approach on the four datasets.}
\begin{center}
\begin{tabular}{lcccc}
\hline
                        & Accuracy (\%)  \\
\hline
UCF-Sports              & 95.7  \\
J-HMDB                  & 67.2       \\
UCF-101 (24 actions)    & 94.4   \\
\hline
\end{tabular}
\label{tab:rec}
\end{center}
\end{table}

\begin{figure*}[!t]
\centering
\includegraphics[width=0.95\linewidth]{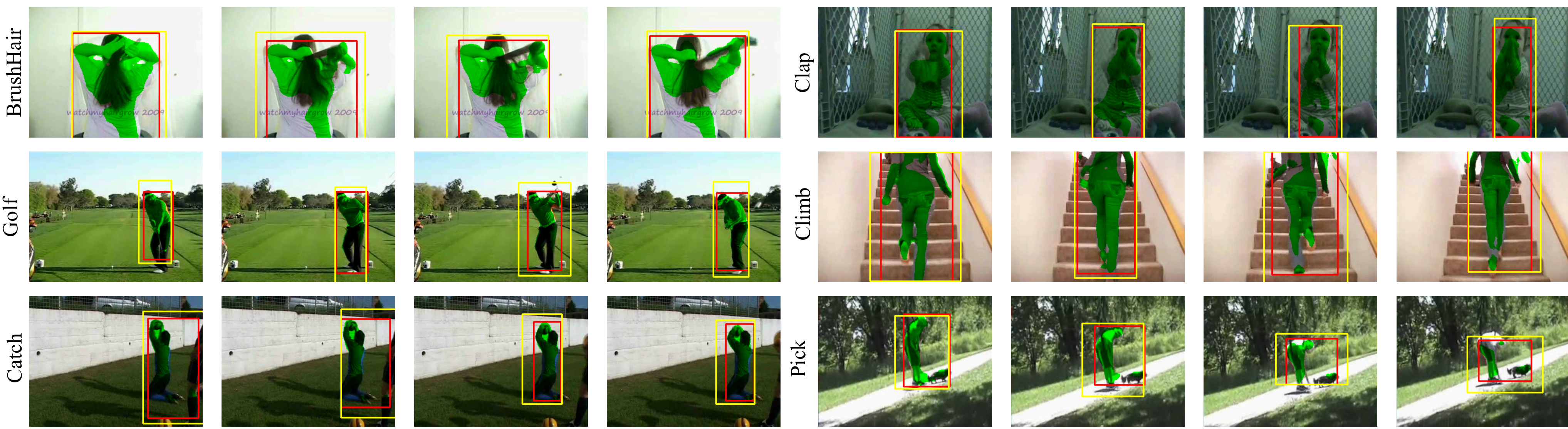}
\caption{Action segmentation and detection results obtained by ST-CNN on the J-HMDB dataset. Green pixel-wise segmentation maps show the predictions of ST-CNN, and the red boxes show the bounding boxes generated from the segmentation maps. Yellow boxes represent the detection results obtained by T-CNN for comparison.}
\label{fig:jhmdb_seg_results}
\end{figure*}

\subsubsection{T-CNN vs. ST-CNN for Action Detection}

We evaluate the proposed ST-CNN method on the J-HMDB dataset using two performance measures, \ie~bounding box action detection and  pixel-wise segmentation maps, since both bounding box and segmentation maps annotations are provided for this dataset. For action detection evaluation, we generate the bounding box for each frame based on the enclosing box of the predicted segmentation map. We report the video mean average precision (mAP) results of our method and several state-of-the-art detection approaches in Table \ref{tab:jhmdb}. ST-CNN achieves the best results and outperforms T-CNN by 0.8\%. We also compare ST-CNN with a baseline method \cite{lu2015human} in terms of action segmentation, and our method performs significantly better than the baseline, leading to almost 10\% improvement of IoU. Some qualitative action segmentation and action detection results of our ST-CNN approach on the J-HMDB dataset are illustrated in Figure \ref{fig:jhmdb_seg_results}. We also show the detection results of T-CNN for comparison. The predicted segmentation maps of ST-CNN are well aligned with the body contours, which results in more accurate bounding boxes than T-CNN (\eg~``brush hair", ``catch" and ``pick" as shown in Figure \ref{fig:jhmdb_seg_results}).
Although the segmentation noise may cause imperfect bounding box predictions (see the last two example frames of ``climb", where incorrect foreground appear on the hand rail), some post-processing techniques such as conditional random fields (CRFs) can be employed to smooth out the segmentation noise to achieve better segmentation maps.

We compare our 3D sub-pixel convolution with the ``un-pooling + convolution", denoted by Ours (un-pool)) in Table~\ref{tab:jhmdb}. We replace our upsample convolution layer with transpose convolution layer and keep other settings unchanged. By using upsample convolution, we achieve about 5\% accuracy gain compared to transpose convolution based approach.

Since the J-HMDB dataset provides the annotations of pixel-wise segmentation maps, ST-CNN can leverage such pixel-wise labeling information to train the model. However, the other action detection datasets may only have bounding box annotations for actions, such as UCF-Sports. How well  the trained ST-CNN model on J-HMDB is able to generalize to other datasets for action detection with no additional bounding box training? To validate its performance, we directly apply the trained ST-CNN on J-HMDB to the UCF-Sports dataset for generating action segmentation maps. The tightest bounding box that covers the foreground segmentation in each frame is considered as the action bounding box prediction. Beside bounding box prediction, since the UCF-Sports dataset consists of different action categories, we need to train the action recognition classifier on UCF-Sports to predict which action the tubes belong to. We follow the same experimental settings as in Section~\ref{subsubsec:ucfsports}, and report the results in Table~\ref{tab:per-class}. We see that although ST-CNN is trained on a different dataset (J-HMDB), it still achieves a frame level mAP of 84.5\%, which is comparable to T-CNN with a mAP of 86.7\%. Note that T-CNN is trained on UCF-Sports for bounding box prediction.
The results show that the ST-CNN model trained on J-HMDB is able to well distinguish between action regions and background, demonstrating its high capacity of generalizing to cross-dataset action detection. Moreover, ST-CNN outperforms other competing action detection approaches.

\begin{figure*}
\centering
\includegraphics[width=0.93\linewidth]{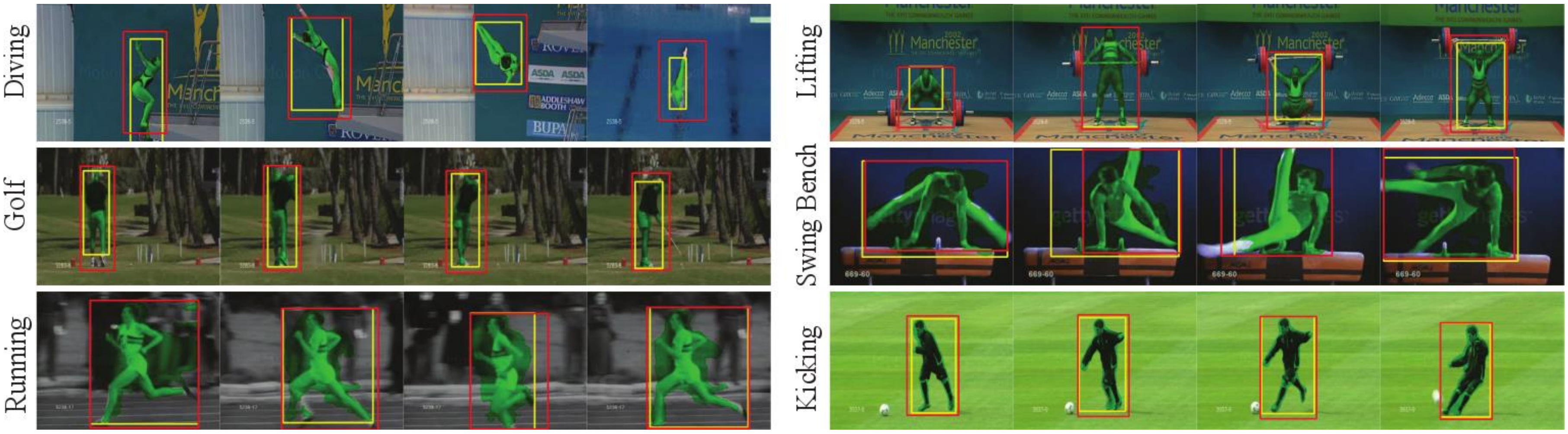}
\caption{Transfer learning from J-HMDB to UCF-Sports. We compare ST-CNN action segmentation (green) and bounding box detection (yellow) with ground truth (red) on selected frames of testing videos.}
\label{fig:sports_mtcnn_example}
\end{figure*}

\subsection{Video Segmentation Experiments}
\label{subsec:seg_exp}
Densely Annotated VIdeo Segmentation (DAVIS) 2016 dataset is specifically designed for the task of video object segmentation. It consists of 50 videos with 3455 annotated frames.
Consistent with most prior work, we conduct experiments on the 480p videos with a resolution of $854 \times 480$ pixels. 30 videos are taken for training and 20 for validation.

For this experiment, we only need the segmentation branch of our ST-CNN network. Before the training process, we gather the videos from three datasets: JHMDB \cite{Jhuang:ICCV:2013}, DAVIS'16 \cite{Perazzi2016} (only training set) and SegTrackv2 \cite{almomani2013segtrack}. We resize all the frames and annotations to  fixed dimensions of  $320\times240$. If there are multiple foreground objects in the annotation, we merge them as a single foreground object and consider other pixels as background. We train the model for $100k$ iterations and fix the learning rate at $10^{-3}$. We call this trained model as our pre-trained model for video segmentation.

We then use this pre-trained model and finetune it on DAVIS'16 training set for $8k$ iterations with learning rate $10^{-4}$ and $2k$ more iterations with learning rate $10^{-5}$. Due to GPU memory constraint, we are not able to process the high resolution videos in 3D CNN. In the training phase, we resize input videos to $320\times240$ and adjust ground truth annotations to the corresponding resolution. During testing, test videos are resized to $320\times240$ and fed into the network. To make our results comparable to ground truth, the $320\times240$ segmentation maps from the network are upscaled to $854\times480$ by bi-linear interpolation. Finally, our segmentation boundaries are smoothed by fully connected conditional random field \cite{krahenbuhl2011efficient}.

Due to limited number of training videos in DAVIS'16 dataset, we also perform data augmentation to increase the number of video clips for training. Specifically, the following data augmentation techniques are considered:
\begin{itemize}[leftmargin=*]
\item \textbf{Illumination Modification.} We translate the video pixels from RGB space to HSV space, since the Saturation (S) channel correspond to ``lightness", while value (V) channel determines the ``brightness". The V channel value $v$ is randomly altered to $v'$ via $v'=av$ where $a\in1\pm0.1$.
\item \textbf{Background replacement with Foreground.} Half of the foreground (either top/down or left/right) is replaced by background. We use the nearest pixels to interpolate the region.
\item \textbf{Clip Flipping and Shifting.} We randomly select clips to flip horizontally, and shift the image by 1 pixel.
\end{itemize}

We adopt the same evaluation setting as reported in \cite{Perazzi2016}. There are three parts. \textbf{Region Similarity} $\mathcal{J}$, which is obtained by IoU between the prediction and the ground-truth segmentation map. \textbf{Contour Accuracy} $\mathcal{F}$ measures the contours accuracy. \textbf{Temporal Stability} $\mathcal{T}$ tracks the temporal consistency in a video. For the first two evaluation, we report the mean, recall and decay. For the third one, we report the average.
We compare our results with several unsupervised implementations, since our approach does not require any manual annotation or prior information about the object to be segmented, which is defined as unsupervised segmentation. It is different from the semi-supervised approaches which assume the ground truth segmentation map of the first frame of a test video is given. Apparently, unsupervised segmentation is a much harder task, but is more practical since it does not require any human labelling effort during testing once the segmentation model has been trained. We compare our method with the state-of-the-art unsupervised approaches in Table~\ref{tab:davis_overall}. According to the results, our method achieves the best performance in all performance metrics. Compared to ARP \cite{koh2017primary}, the previous state-of-the-art unsupervised approach, our method achieves 5\% gain in contour accuracy ($\mathcal{F}$) and 15\% gain in temporal stability ($\mathcal{T}$), demonstrating that 3D CNN can effectively take advantage of the temporal information in video frames to achieve temporal segmentation consistency.

\begin{figure*}[!bt]
\small
\centering

\begin{tikzpicture}
\begin{axis}[
ybar,
enlargelimits=0.015,
bar width=2pt,
width=\linewidth,
height=0.24\linewidth,
x tick label style={rotate=30, anchor=east},
xticklabels={Blackswan, Bmx-Trees, Breakdance, Camel,
Car-Roundabout, Car-Shadow, Cows, Dance-Twirl, Dog,
Drift-Chicane, Drift-Straight, Goat, Horsejump-High,
Kite-Surf, Libby, Motocross-Jump, Paragliding-Launch,
Parkour, Scooter-Black, Soapbox},
xtick=data,
ylabel=Mean Jaccard index ($\mathcal{J}$),
	legend style={at={(0.76,1.04)}, anchor=north,legend columns=-1},
]

\addplot coordinates {(0, 92.3) (1, 52.9) (2, 66.9) (3, 86.8) (4, 92.1) (5, 91.6) (6, 89.9) (7, 72.2) (8, 86.3) (9, 77.5) (10, 81.6) (11, 79.3) (12, 74.0) (13, 65.3) (14, 66.8) (15, 79.6) (16, 53.2) (17, 78.1) (18, 80.8) (19, 85.0)};
\addplot coordinates {(0, 88.1) (1, 49.9) (2, 76.2) (3, 90.3) (4, 81.6) (5, 73.6) (6, 90.8) (7, 79.8) (8, 71.8) (9, 79.7) (10, 71.5) (11, 77.6) (12, 83.8) (13, 59.1) (14, 65.4) (15, 82.3) (16, 60.1) (17, 82.8) (18, 74.6) (19, 84.6)};
\addplot coordinates {(0, 74.1) (1, 43.3) (2, 37.1) (3, 88.1) (4, 88.6) (5, 92.0) (6, 90.2) (7, 81.0) (8, 88.7) (9, 63.9) (10, 84.9) (11, 82.3) (12, 82.4) (13, 64.6) (14, 69.0) (15, 80.5) (16, 62.2) (17, 84.9) (18, 71.8) (19, 81.3)};
\addplot coordinates {(0, 81.2) (1, 52.2) (2, 51.2) (3, 83.6) (4, 90.7) (5, 89.6) (6, 86.9) (7, 70.4) (8, 88.9) (9, 59.6) (10, 81.1) (11, 83.0) (12, 65.2) (13, 39.2) (14, 58.4) (15, 77.5) (16, 57.1) (17, 76.0) (18, 68.8) (19, 62.4)};


\legend{Ours,ARP \cite{koh2017primary}, LVO \cite{tokmakov2017learning},FSEG \cite{jain2017fusionseg}}
\end{axis}
\end{tikzpicture}
\caption{Comparison of Mean Jaccard index ($\mathcal{J}$) of different approaches on each of the sequences independently.}
\label{fig:seg_ind}
\end{figure*}
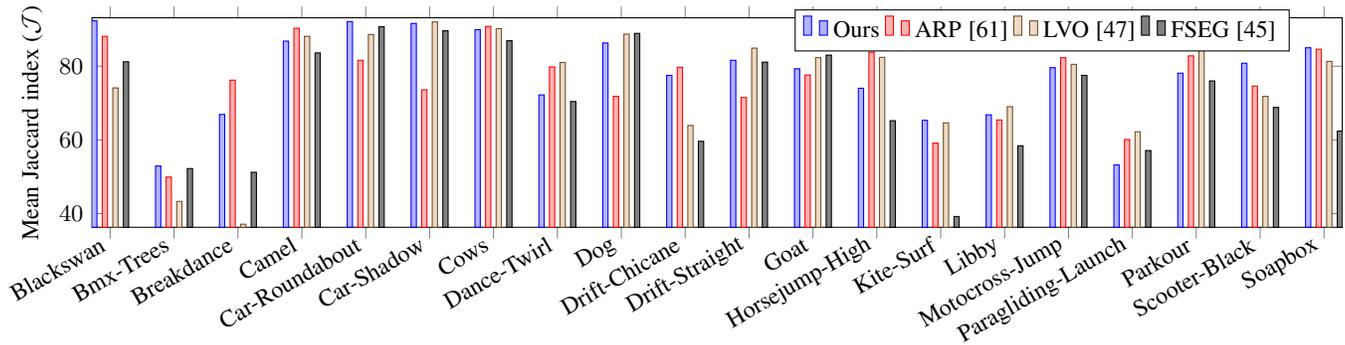

\begin{figure*}
\centering
\includegraphics[width=0.93\linewidth]{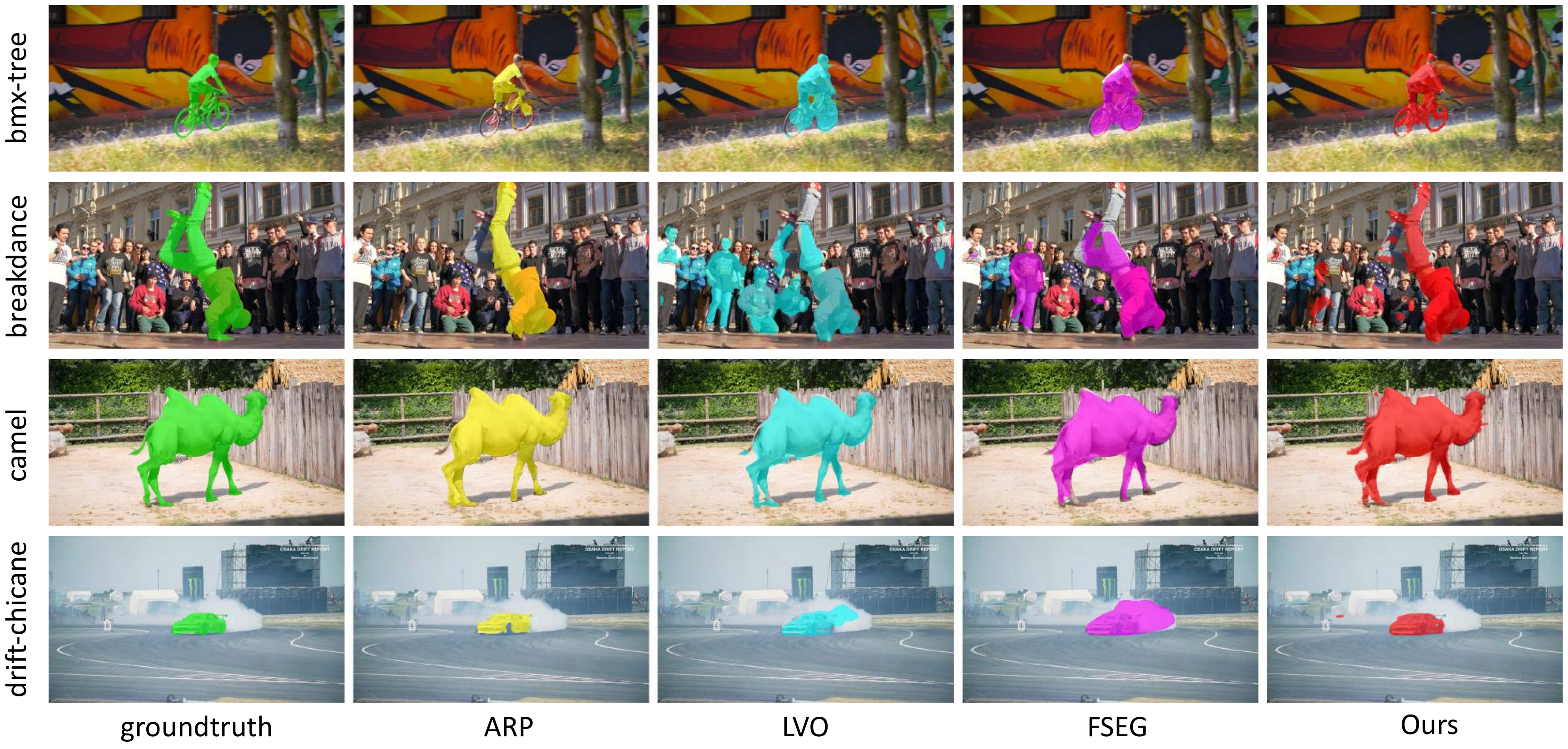}
\caption{Qualitative results of the proposed approach (red), ARP (yellow), LVO (cyan) and FSEG (magenta) on selected frames from DAVIS dataset.}
\label{fig:mtcnn_example}
\end{figure*}

Figure~\ref{fig:seg_ind} shows the quantitative results per video sequence of our approach and the next three top performing methods on DAVIS dataset: ARP \cite{koh2017primary}, LVO \cite{tokmakov2017learning} and FSEG \cite{jain2017fusionseg}. Our approach performs the best on low contrast videos including Blackswan, Car-Roundabout and Scooter-Black and achieves competitive results on other videos. Figure~\ref{fig:mtcnn_example} presents the qualitative results on four video sequences. In the first row, our results are the most accurate. Our method is the only one which can detect the rims. In the second row, ARP performs the best in suppressing background. However, only our approach detects both legs. The third row shows that only our method is able to accurately segment the tail of the camel. The last row is a very challenging video because of the smoke and small initial size of the car. ARP misses part of the car, while LVO and FSEG mis-classify part of background as moving object. However, our method segments out the car completely and accurately from the background smoke in the scene.

\begin{table*}
\caption{Overall results of region similarity ($\mathcal{J}$), contour accuracy ($\mathcal{F}$) and temporal stability ($\mathcal{T}$) for different approaches. $\uparrow$ means the more the better and $\downarrow$ means the less the better.}
\centering
\small
\begin{tabular}{c|c|p{0.8cm}p{0.8cm}p{0.8cm}p{0.8cm}p{0.8cm}p{0.8cm}p{0.8cm}p{0.8cm}p{0.8cm}p{0.8cm}p{0.8cm}p{0.8cm}}

\hline

\multicolumn{2}{c|}{Measure}            & ARP \cite{koh2017primary}   & FSEG \cite{jain2017fusionseg}  & LMP \cite{tokmakov2016learning}    & FST \cite{papazoglou2013fast}   & CUT \cite{keuper2015motion}   & NLC \cite{faktor2014video}    & MSG \cite{brox2010object}   & KEY \cite{lee2011key}   & CVOS \cite{taylor2015causal} & TRC \cite{fragkiadaki2012video}    & SAL \cite{wang2015saliency} & \textbf{Ours}\\
\hline
\multirow{3}{*}{$\mathcal{J}$}  & Mean $\uparrow$  & 76.2	& 70.7	& 70.0	& 55.8	& 55.2	& 55.1	& 53.3	& 49.8	& 48.2	& 47.3	& 39.3  & \textbf{77.6} \\
                                & Recall $\uparrow$& 91.1	& 83.5	& 85.0	& 64.9	& 57.5	& 55.8	& 61.6	& 59.1	& 54.0	& 49.3	& 30.0  & \textbf{95.2} \\
                                & Decay $\downarrow$& 7.0	& 1.5	& 1.3	& \textbf{0.0}	& 2.2   & 12.6	& 2.4	& 14.1	& 10.5	& 8.3	& 6.9   & 2.3 \\
\hline
\multirow{3}{*}{$\mathcal{F}$}  & Mean $\uparrow$  & 70.6	& 65.3	& 65.9	& 51.1	& 55.2	& 52.3	& 50.8	& 42.7	& 44.7	& 44.1	& 34.4  & \textbf{75.5} \\
                                & Recall $\uparrow$& 83.5  & 73.8	& 79.2	& 51.6	& 61.0	& 51.9	& 60.0	& 37.5	& 52.6	& 43.6	& 15.4  & \textbf{94.7} \\
                                & Decay $\downarrow$ & 7.9   & \textbf{1.8}	& 2.5	& 2.9	& 3.4	& 11.4	& 5.1	& 10.6	& 11.7	& 12.9	& 4.3   & 4.9 \\
\hline
$\mathcal{T}$                   & Mean $\downarrow$  & 39.3	& 32.8	& 57.2	& 36.6	& 27.7	& 42.5	& 30.1	& 26.9	& 25.0	& 39.1	& 66.1  & \textbf{22.0} \\
\hline
\end{tabular}
\label{tab:davis_overall}
\end{table*}

\subsection{Computational Cost}
We carry out our experiments on a workstation with one GPU (Nvidia GTX Titan X). Given a 40-frames video, T-CNN pipeline takes 1.1 seconds to generate tube proposals, 0.03 seconds to link tube proposals in a video and 0.9 seconds to predict action label.
ST-CNN takes only 0.7 seconds to detect actions (including about 0.6 seconds for segmentation), which is 3 times faster than T-CNN. In ST-CNN, the video clips are only fed into the network once and the detection results are obtained, while in T-CNN, the clips are fed into TPN first to get the tube proposals. Then both clips and tubes are used in the recognition network to predict the label. Since the input to the first stage are clips with fixed duration, while the input to the second stage is whole videos with various duration, it is hard to share weights in their common layers. In contrast, in ST-CNN, there is only one stage since we avoid the proposal generation part. Therefore, ST-CNN is more efficient than T-CNN.

\section{Conclusion}
\label{sec:conclusion}
In this paper we propose an end-to-end 3D CNN based pipeline for action detection in videos. It exploits 3D CNN to extract effective spatio-temporal features and perform action segmentation and recognition. We explored two approaches to locate the action. In the first, coarse proposal boxes are densely sampled based on the 3D convolutional feature cube and linked together (T-CNN). In the second, pixels in each frame are segmented into foreground/background and foreground pixels are aggregated into action segments (ST-CNN). Extensive experiments on several benchmark datasets demonstrate the strength of our approach for spatio-temporal action detection as well as segmentation compared to the state-of-the-arts.

\ifCLASSOPTIONcompsoc
  \section*{Acknowledgments}
\else
  \section*{Acknowledgment}
\fi

The project was supported by Award No.
2015-R2-CX-K025, awarded by the National Institute of Justice,
Office of Justice Programs, U.S. Department of Justice. The opinions,
findings, and conclusions or recommendations expressed in
this publication are those of the author(s) and do not necessarily
reflect those of the Department of Justice.

\ifCLASSOPTIONcaptionsoff
  \newpage
\fi

\bibliographystyle{IEEEtran}
\bibliography{egbib}

\begin{IEEEbiography}[{\includegraphics[width=1in,height=1.25in,clip,keepaspectratio]{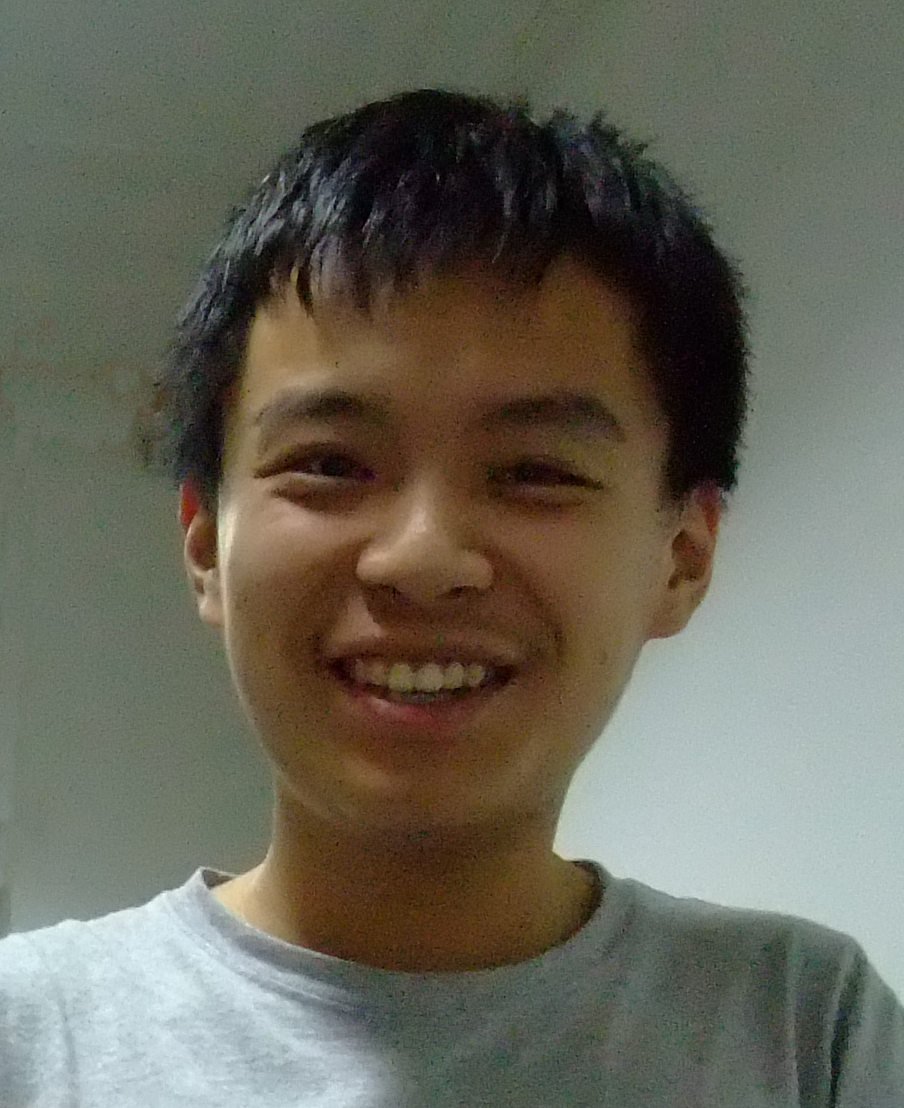}}]{Rui Hou}
received his BSc from Beijing University of Posts and Telecommunications in 2010 and MS degree in Electrical Engineering from Arizona State University. He joined Center for Research in Computer Vision at University of Central Florida in 2012, where he is currently working toward the PhD degree in Computer Science. His research interests include action recognition/detection in video and semantic segmentation in video.
\end{IEEEbiography}

\vspace{-4.25em}

\begin{IEEEbiography}[{\includegraphics[width=1in,height=1.25in,clip,keepaspectratio]{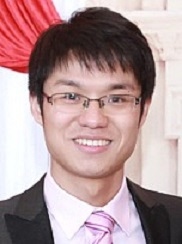}}]{Chen Chen}
received the PhD degree in the Department of Electrical Engineering at the University of Texas at Dallas in 2016. He is currently a Post-Doc in the Center for Research in Computer Vision at University of Central Florida. His research interests include compressed sensing, signal and image processing, and computer vision. He is an associated editor for KSII Transactions on Internet and Information Systems, and Signal, Image and Video Processing. 
\end{IEEEbiography}


\vspace{-4.25em}

\begin{IEEEbiography}[{\includegraphics[width=1in,height=1.25in,clip,keepaspectratio]{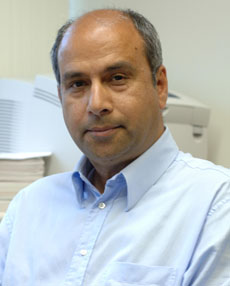}}]{Mubarak Shah,} the Trustee chair professor of computer science, is the founding director of the Center for Research in Computer Vision at the University of Central Florida. He is a fellow of IEEE, AAAS, IAPR and SPIE. He is an editor of an international book series on video computing, editor-in-chief of Machine Vision and Applications journal, and an associate editor of ACM Computing Surveys journal. He was the program cochair of the CVPR in 2008, an associate editor of the IEEE TPAMI.

\end{IEEEbiography}




\end{document}